\documentclass[10pt,twocolumn,letterpaper]{article}

\usepackage{iccv}
\usepackage{times}
\usepackage{epsfig}

\usepackage{graphicx}
\usepackage{amsmath}
\usepackage{amsfonts}
\usepackage{amssymb}
\usepackage{booktabs}
\usepackage{multirow}
\usepackage{float}
\usepackage{paralist}
\usepackage{caption, overpic, textpos}
\usepackage{bm}
\usepackage{caption}
\usepackage{subcaption}

\newtheorem{theorem}{Theorem}
\newtheorem{definition}{Definition}
\newtheorem{example}{Example}
\newtheorem{proof}{Proof}

\usepackage[pagebackref=true,breaklinks=true,letterpaper=true,colorlinks,bookmarks=false]{hyperref}

\iccvfinalcopy 

\def\iccvPaperID{9562} 

\newcommand{\bx}{\bm{x}}

\newcommand{\hy}{\hat{y}}

\begin{document}

\title{SEAL: Simultaneous Label Hierarchy Exploration And Learning}

\author{Zhiquan Tan\thanks{Equal Contribution}\\
Department of Mathematical Sciences\\
Tsinghua University\\
{\tt\small tanzq21@mails.tsinghua.edu.cn}
\and
Zihao Wang$^*$\\
Department of CSE\\
HKUST\\
{\tt\small zwanggc@cse.ust.hk}
\and 
Yifan Zhang$^*$\\
IIIS\\
Tsinghua Universtiy\\
{\tt\small zhangyif21@mails.tsinghua.edu.cn}
}

\maketitle
\ificcvfinal\thispagestyle{empty}\fi

\begin{abstract}

Label hierarchy is an important source of external knowledge that can enhance classification performance. However, most existing methods rely on predefined label hierarchies that may not match the data distribution. To address this issue, we propose \textbf{S}imultaneous label hierarchy \textbf{E}xploration \textbf{A}nd \textbf{L}earning (SEAL), a new framework that explores the label hierarchy by augmenting the observed labels with latent labels that follow a prior hierarchical structure. Our approach uses a 1-Wasserstein metric over the tree metric space as an objective function, which enables us to simultaneously learn a data-driven label hierarchy and perform (semi-)supervised learning. We evaluate our method on several datasets and show that it achieves superior results in both supervised and semi-supervised scenarios and reveals insightful label structures. Our implementation is available at \href{https://github.com/tzq1999/SEAL}{https://github.com/tzq1999/SEAL}.

\end{abstract}

\section{Introduction}

Labels play a crucial role in machine learning. They provide the supervision signal for learning models from annotated data. However, obtaining label annotations is often costly and time-consuming, which motivates the study of semi-supervised learning that leverages both labeled and unlabeled data~\cite{chen2020big,assran2021semi,wang2021data}. A common technique for semi-supervised learning is also related to the label, specifically, using (pseudo-)labels~\cite{berthelot2019mixmatch,sohn2020fixmatch,gong2021alphamatch}.  Unlabeled data is augmented in different ways~\cite{xie2020unsupervised}, and pseudo labels are then generated from model predictions for different augmentations of the same data. The model is updated by enforcing the consistency of pseudo labels across augmentations. This technique is known as ``consistency regularization’'~\cite{Rasmus2015SemiSupervisedLW}.

Labels are also important for understanding data, as they link real-world observations with abstract semantics. It has been shown that exploiting hierarchical structures of label semantics can enhance the performance of supervised and semi-supervised learning. These structures can be obtained from external sources such as decision trees~\cite{wan2020nbdt} and knowledge graphs~\cite{miller1998wordnet,speer2017conceptnet}.
Once the label hierarchy is available, models can be trained by either (1) predicting hierarchical semantic embeddings jointly with labels~\cite{deng2010does,deng2012hedging,barz2019hierarchy,liu2020hyperbolic,wang2021hierarchical,karthik2021no,nassar2021all} or (2) optimizing hierarchical objective functions that incorporate label relations~\cite{bertinetto2020making,bilal2017convolutional,garnot2020leveraging,garg2022learning}. Alternatively, the structure can also be incorporated into the model architecture itself~\cite{kontschieder2015deep,wu2016learning,wan2020nbdt,chang2021your,garg2022hiermatch}.

\begin{figure*}[t] 
\centering 
\includegraphics[width=1.8\columnwidth]{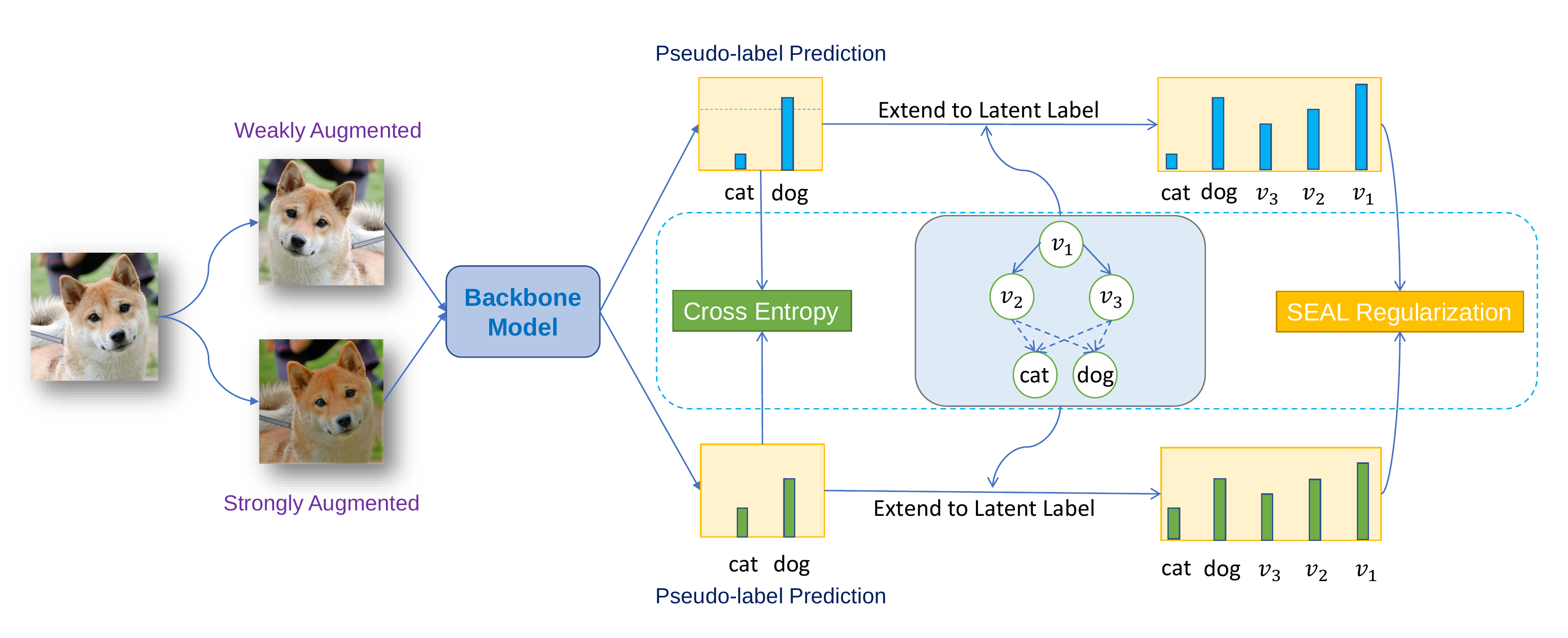}
\caption{An illustration of the SEAL framework. After getting the predicted probability of a weakly augmented image on \emph{observed} labels, the model shall use any pseudo labeling techniques to give it a pseudo label. Then the cross entropy loss between the pseudo label and predicted probability on the strong-augmented image will comprise part of the SEAL regularization. Given a (soft) tree hierarchy can introduce the \emph{latent} labels. Extending the predicted probability on strong-augmented image and pseudo label to \emph{total} labels will give a SEAL regularization loss (relaxed Tree-Wasserstein loss) which is another part of the SEAL regularization. Note that the backbone model and the SEAL regularization model (label hierarchies) can be updated via gradient descent simultaneously.}
\label{arch}
\end{figure*}

Although predefined label hierarchies are frequently used in the existing literature, they cannot always match the actual data distribution. However, not much effort has been made to derive the label hierarchy from a data-driven perspective. To address this issue, we propose 
\textbf{S}imultaneous label hierarchy \textbf{E}xploration \textbf{A}nd \textbf{L}earning (SEAL), which achieves two goals by incorporating an additional regularization term.

The first goal of SEAL is to identify the data-driven label hierarchy. This goal differs from hierarchical clustering, which discovers hierarchical structures from data that do not align with labels. SEAL expands the label alphabet by adding unobserved \textit{latent} labels to the \textit{observed} label alphabet. The data-driven label hierarchy is modeled by combining the predefined hierarchical structure of \textit{latent} labels with the optimizable assignment between observed and latent labels. The observation that inspires this approach is that the labels are typically a subset of concepts in a larger knowledge graph~\cite{miller1998wordnet}.

The second goal of SEAL is to improve the classification performance of state-of-the-art methods. To accomplish this goal, we propose a new regularization term that uses the model predictions and the label annotations on the \textit{observed} label alphabet to encourage agreement on both \textit{observed} and \textit{latent} label alphabets. The confidence on \textit{latent} labels is estimated by simulating the Markov chain based on the label hierarchy between \textit{observed} and \textit{latent} labels. This regularization term can be added to all existing approaches because of the universality of comparing the prediction and labels on the \textit{observed} label alphabet.

SEAL's soundness and effectiveness are validated theoretically and empirically. The regularization term can be interpreted as a relaxation of the tree Wasserstein metric~\cite{le2019tree}, and it can be used for optimization. Empirical evaluation demonstrates that adding the SEAL framework consistently and significantly improves the classification performance on supervised learning and various semi-supervised learning methods~\cite{sohn2020fixmatch,zhang2021flexmatch,wang2022debiased}. SEAL also achieved a new state-of-the-art for semi-supervised learning on standard datasets. Additionally, our case study shows that the alignment between \textit{observed} and \textit{latent} labels also yields a meaningful label hierarchy.

\section{Related Work}

\subsection{Semi-supervised Learning}


Most existing methods for consistency regularization aim to improve the quality of the pseudo labels generated from unlabeled data. For example, SimPLE~\cite{hu2021simple} introduces a paired loss that minimizes the statistical distance between confident and similar pseudo labels. Dash~\cite{xu2021dash} and FlexMatch~\cite{zhang2021flexmatch} propose dynamic and adaptive strategies for filtering out unreliable pseudo labels during training. MaxMatch~\cite{li2022maxmatch} proposes a worst-case consistency regularization technique that minimizes the maximum inconsistency between an original unlabeled sample and its multiple augmentations with theoretical guarantees. A notable exception is SemCo~\cite{nassar2021all}, which leverages external label semantics to prevent the deterioration of pseudo label quality for visually similar classes in a co-training framework. 

Though proposed in different techniques, all these methods rely on a fixed objective function to define the consistency, which is usually the cross-entropy over the label space. Our work differs from these methods by proposing a novel way to extend the cross-entropy with latent labels and hierarchical structures. Therefore, our method can complement existing methods whenever cross-entropy is used.

\subsection{Label hierarchies}

Label relationships are essential prior knowledge for improving model performance, and can be represented by semantic structures among labels. One prominent form of label structure is the label hierarchy~\cite{garnot2020leveraging}, which can be obtained from external sources like decision trees~\cite{wan2020nbdt} and knowledge graphs~\cite{miller1998wordnet,speer2017conceptnet}. This information can be leveraged to train models as semantic embeddings~\cite{deng2010does, deng2012hedging,barz2019hierarchy,liu2020hyperbolic,wang2021hierarchical,karthik2021no,nassar2021all} or objective functions~\cite{bertinetto2020making, bilal2017convolutional,garnot2020leveraging, garg2022learning}. Additionally, the hierarchical information can also be incorporated as part of the model structure~\cite{kontschieder2015deep,wu2016learning,wan2020nbdt,chang2021your,garg2022hiermatch}.

Pre-defined label hierarchies are widely acknowledged as an essential source of prior knowledge for the label space in classification. This has been extensively discussed in the literature. The label hierarchy information can be used to improve the model training process, such as by embedding the hierarchical labels to maximize the similarity between the latent embedding of the input image and the embedding of its label~\cite{bengio2010label,deng2012hedging,frome2013devise}. This idea has been generalized to various embedding spaces~\cite{barz2019hierarchy,liu2020hyperbolic,garnot2020leveraging} and the joint learning scheme where image embeddings and label embeddings are both optimized~\cite{wu2016learning,chang2021your}. Additionally, hierarchical structures can also be explicitly used to make the training process hierarchical-aware~\cite{deng2010does,bilal2017convolutional,bertinetto2020making,karthik2021no,garg2022hiermatch,garg2022learning}.

However, existing work typically treats the hierarchical label structure as prior knowledge. In contrast, our approach leverages the posterior latent label structures given the presence of labeled and unlabeled samples.

\section{Background}\label{sec:Preliminaries}

In this section, we introduce the notations used throughout this paper first and then define the Tree-Wasserstein distance, which we use to define label hierarchy metrics in Section~\ref{sec:VLL}.

\subsection{Notations}

We consider supervised learning, where $D = \{ (\bx_i, y_i) \}_{i=1}^n$ is the labeled training set. Here, $x_i$ is an image, and $y_i$ is the corresponding (class) label in a set $\mathcal{O}$, where $\mid \mathcal{O} \mid = K$. Without further justification, $y$ is the categorical label and $\delta_y$ is the one-hot label. Our goal is to learn a backbone model $f_\theta$ parameterized by $\theta$, which maps each image $\bx_i$ to a probability over $\mathcal{O}$. We denote the predicted probability vector of image $\bx$ as $f_\theta(\bx)$ and define the pseudo label of $\bx$ as $\hy = \operatorname{argmax}_j f(\bx)^T e_j = \operatorname{argmax}_j \Pr(j|\bx)$, where $e_j$ is the coordinate vector. The objective of supervised learning is to make the ground truth label $y$ and pseudo label $\hy$ consistent.

We also consider semi-supervised learning, where an additional unlabeled dataset $D^{u}$ is provided. During training, we define the relative ratio $\mu$ as the number of unlabeled data to labeled data in a mini-batch. Following RandAugment\cite{cubukpractical}, we shall define weak augmentation $P_{\text{wAug}}(\bx)$ and strong augmentation $P_{\text{sAug}}(\bx)$. These two augmentations are probability distributions of the views of $\bx$ derived by argumentation while keeping the same pseudo label, the term weak and strong describes the distortion density. Moreover, $\bx'$ and $\bx''$ denote the argument data and $y'$ denotes the pseudo-label.

\subsection{Tree-Wasserstein distance}

The Tree-Wasserstein distance~\cite{le2019tree} is a 1-Wasserstein metric~\cite{peyre2019computational} on a tree metric space $(\mathcal{X}, E, w)$, where $\mathcal{X}$ is the node set of a directed rooted tree, $E$ is the (weighted) edge set, and $w=(w_e)_{e \in E}$ denotes the weights of edges $e\in E$. The tree metric $d_{\mathcal{X}}$ between any two nodes of a tree is defined as the length of the shortest path between them. Given node $v \in \mathcal{X}$, let $\Gamma(v)$ be the set of nodes in the subtree of $\mathcal{X}$ whose root node is $v$. For each weighted edge $e\in E$, we denote the deeper (away from the root) level endpoints of weighted edge $e$ as $v_e$. Then, the Tree-Wasserstein metric can be computed in closed form, as shown in Theorem~\ref{thm:tw}.

\begin{theorem}[Tree-Wasserstein distance \cite{le2019tree}]~\label{thm:tw}
Given two probability measures $\mu, \nu$ supported on a directed rooted tree $\mathcal{X}$, and choosing the ground metric as tree metric $d_{\mathcal{X}}$, then the Wasserstein-$1$ distance under Definition \ref{wass def} can be reformulated as follows:
\begin{align}
W_{d_\mathcal{X}}(\mu, \nu)=\sum_{e\in E} w_e\left|\mu\left(\Gamma\left(v_e\right)\right)-\nu\left(\Gamma\left(v_e\right)\right)\right|.    
\end{align}
\end{theorem}

\section{The SEAL framework}\label{sec:VLL}

We propose a novel framework called Simultaneous label hierarchy Exploration and Learning (SEAL), which is motivated by probabilistic models that leverage latent structures to improve classification performance. In the SEAL framework, we introduce \textit{latent} labels that are aligned with \textit{observed} labels to capture the data-driven label hierarchy.

To incorporate the latent labels into the learning process, we propose SEAL regularization, which extends the loss function from the \textit{observed} label alphabet to the \textit{total} label alphabet. This regularization encourages the agreement between the model predictions and both the observed and latent label alphabets, improving the classification performance of state-of-the-art methods.

Finally, we demonstrate that SEAL regularization enables simultaneous label hierarchy exploration and learning. By leveraging the label hierarchy between the observed and latent labels, SEAL can learn a more meaningful and accurate label hierarchy that aligns with the data distribution. The effectiveness and soundness of our proposed SEAL framework are validated both theoretically and empirically.

\subsection{Why latent structure?}

\begin{example}\label{eg:motivating}
    Let us consider a classifier that predicts the probability of an image belonging whether an image belongs to the "apple" or "paint" class. Although the two classes may seem unrelated, they may share some hidden structures, such as the colors "red" and "green". Knowing the conditional probabilities of colors given each class, we can calculate the probability of an image being red given its probability of belonging to the "apple" class. Suppose we know the conditional probabilities $\Pr(\text{red}|\text{apple})=0.9$, $\Pr(\text{green}|\text{apple})=0.1$, $\Pr(\text{red}|\text{paint})=0.5$, and $\Pr(\text{green}|\text{paint})=0.5$. In this scenario, if the classifier assigns a probability of $0.8$ for an image being an apple, the question arises as to what is the probability that this image is red. By applying the law of probability, the answer is $0.8 \times 0.9 + 0.2 \times 0.5 = 0.82$.
    \begin{figure}[t]
    \centering 
    \includegraphics[width=1\columnwidth]{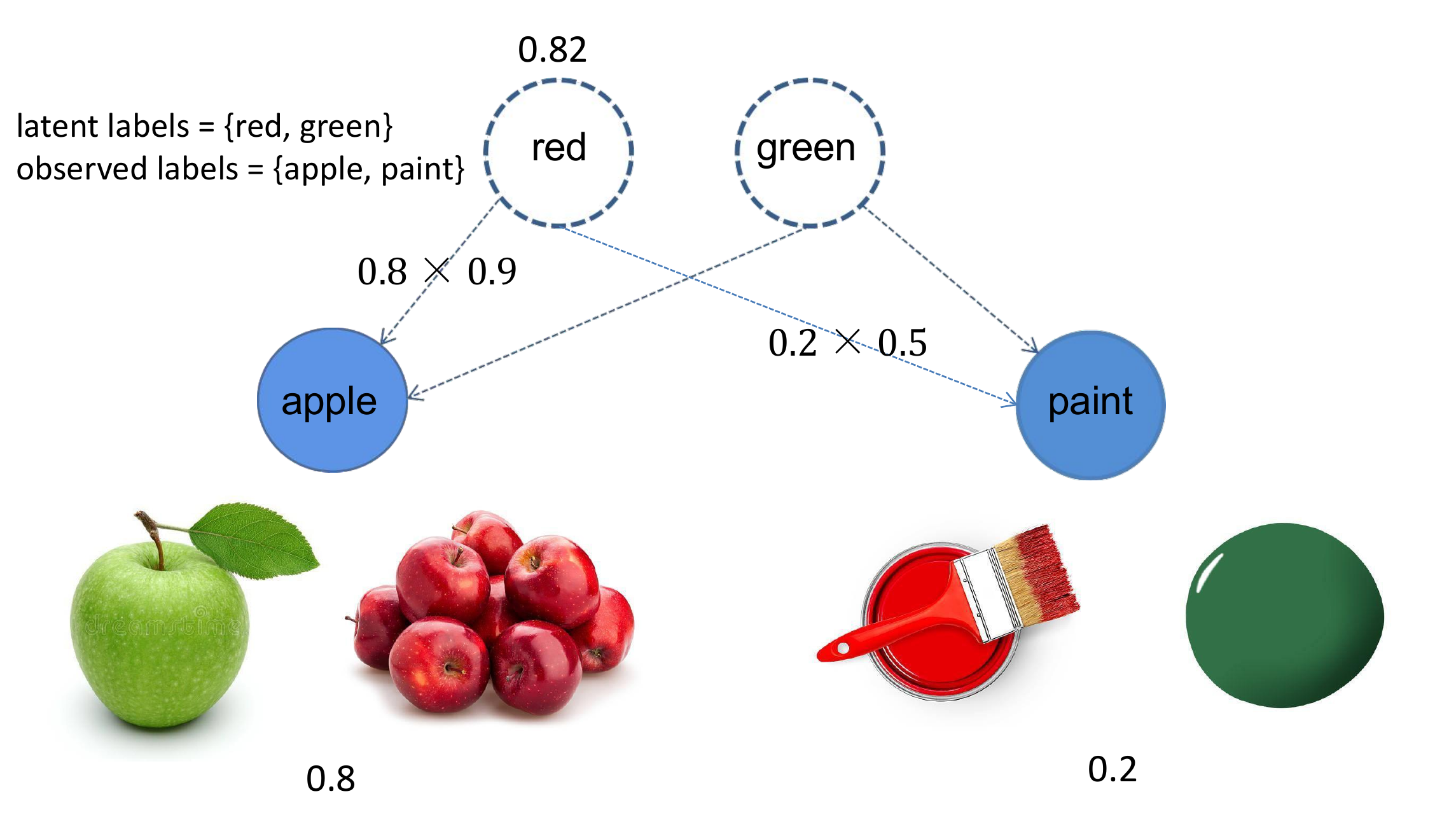} 
    \label{fig:graph-illustration} 
    \caption{An illustrative example}
    \end{figure}
\end{example}

\subsection{Latent structure and SEAL regularization}

The Example~\ref{eg:motivating} illustrates that considering latent variables, such as color in this case, can provide more information to the classifier, leading to better performance. In the context of image classification, these latent variables can represent various factors, such as textures, shapes, and semantic meanings. However, identifying these latent variables and modeling their relationships with observed variables is not always straightforward, which is the focus of the SEAL framework. 

Then we formally present the following definitions below. The set of labels $\mathcal{O}$ in the dataset is denoted as \textit{observed} label alphabets. The set of latent labels $\mathcal{L}$ is denoted as \textit{latent} label alphabet. We call $\mathcal{O}\cup \mathcal{L}$ \textit{total} label alphabet. Let $|\mathcal{O}| = K$ and $|\mathcal{O}\cup \mathcal{L}| = N$ be the sizes of observed and total label alphabets.

The relationship between \textit{observed} and \textit{latent} labels are described by (directed) graphs. Specifically, let $A_1$ be the adjacency matrix for latent labels in $\mathcal{L}$ and $A_2$ be the connection matrix between $\mathcal{L}$ and $\mathcal{O}$. The adjacency matrix $A$ for $\mathcal{L}\cup \mathcal{O}$ characterizes the latent structure.
\begin{align}
     A=\left(\begin{array}{cc}
{A}_{1} & {A}_{2} \\
{0} & {0}
\end{array}\right).
\end{align}
It is assumed to no connection exists inside $\mathcal{O}$. 
SEAL is targeted to discover the hierarchical structure of labels, therefore, additional assumptions are imposed on matrix $A$.
The key assumption of $A$ follows the following theorem.
\begin{theorem}[\cite{takezawa2021supervised}]\label{thm:tree-assumption} Suppose a directed graph $G$ having a total of $N$ nodes, which we denote as $\{ v_1, v_2, . . . , v_N \}$. If the adjacency matrix $A \in\{0,1\}^{N \times N}$ of this graph satisfies the following conditions:
\begin{enumerate}
    \item $A$ is a strictly upper triangular matrix.
    \item $A^T {1}_{N}=(0,1, \cdots, 1)^{\top}$.
\end{enumerate}
then $G$ is a directed rooted tree with $v_{1}$ as the root.
\end{theorem}
The hierarchical structure of observed labels is then described by the graph defined by $A$, satisfying conditions in Theorem~\ref{thm:tree-assumption}.

We introduce a weight matrix $\alpha_{sr}\in\mathbb{R}^{(N-K)\times K}$ that describes the connection from $s\in\mathcal{L}$ to $r\in\mathcal{O}$ to further quantify how much an observed label contributes to a latent label. A \textbf{SEAL extension} is then defined by a five-tuple $(\mathcal{O}, \mathcal{L}, A_1, A_2, \alpha)$. We note that $\alpha$ should be related to the $A_2$ and its specific formulation will be detailed in the following parts.

Then, we are able to extend the model's prediction on the observed label alphabet $\mathcal{O}$ to the total label alphabet.
\begin{definition}[Total prediction and total target]
Let $p_r$ be the probability of the label $r\in \mathcal{O}$, vector $q$ on \emph{total} label alphabet $\mathcal{O}\cup \mathcal{L}$ is
\begin{equation}
[q(\mu)]_s= \begin{cases} \mu_s & \text { if } v \in \mathcal{O} \\ \sum_{r\in \mathcal{O} }\alpha_{sr} \mu_r & \text { if }v \in \mathcal{L} \end{cases},
\end{equation}
where $s \in \mathcal{O} \cup \mathcal{L}$. Given a sample of input and (pseudo-)label $(\bx, y)$, we note that the $p_r$ could be derived by both the model prediction $f_\theta(\bx)$, the one-hot label $\delta_y$, or the pseudo-label $\delta_{y'}$.
Moreover, $q(f_\theta(\bx))$ is the total prediction while $q(\delta_y)$ is denoted as the total target.
\end{definition}

We note that $q$ is not the probability in any case, since it extends the original probability $p$ over $\mathcal{O}$ by further considering the aggregations over $\mathcal{L}$. However, it is also sufficient to define objective functions to minimize the differences between total prediction and total target, which is SEAL regularization.
\begin{definition}[SEAL regularization]
    Given input $\bx$, target $y$, model $f_\theta$, and a SEAL extension $(\mathcal{O}, \mathcal{L}, A_1, A_2, \alpha)$, the SEAL regularization is defined as $\phi(f_\theta(\bx),\delta_y) = D(q(f_\theta(\bx)),  q(\delta_y))$, where $D$ is a distance function. 
\end{definition}

In this paper, we consider SEAL regularization where $D$ is the weighted $\ell_1$ metric:
\begin{align}
    & \phi(f_\theta(\bx),\delta_y) = D(q(f_\theta(\bx)), q(\delta_y)) \nonumber\\
    = &\sum_{s\in \mathcal{O}\cup \mathcal{L}} w_s | [q(f_\theta(\bx))]_s - [q(\delta_y)]_s |,
\end{align}
where $w_s$ is the weight for each observed or latent label.

We have presented the basic framework of SEAL, then we detail how SEAL is used to explore label hierarchy and improve learning in the next parts.

\subsection{Label Hierarchy Exploration with SEAL}

In this section, we explore the label hierarchy under the SEAL extension $(\mathcal{O}, \mathcal{L}, A_1, A_2, \alpha)$. To achieve this, we first specify $\mathcal{L}$, $A_1$, $A_2$, and $\alpha$, which breakdowns into two tasks. The first task is to specify $\mathcal{L}$ and $A_1$ to define the prior structure inside $\mathcal{L}$, while the second task is to specify how $\alpha$ and $A_2$ are related to defining how the structure is optimized.

\paragraph{Task (a): Prior Structure Specification.}
To specify the prior structure inside $\mathcal{L}$, we choose $A_1$ to be a trivial binary tree or trees derived from prior knowledge such as a part of a knowledge graph or a decision tree. This choice of $A_1$ allows us to control the prior structure of the label hierarchy and incorporate prior domain knowledge. Additionally, we can use the hierarchical structure of $A_1$ to guide the training of the model to improve performance.

\paragraph{Task (b): Structure Optimization Specification.}
To specify how $\alpha$ and $A_2$ are related to defining how the structure is optimized, we note that $\alpha$ and $A_2$ both reflect how the structure interacts with the model and data. Specifically, we compute $\alpha$ from $A_2$ from the Markov chain on trees. This choice of $\alpha$ emphasizes more on the prediction of the model while $A_2$ emphasizes more on the interpretation of the label hierarchy.

In summary, by specifying $\mathcal{L}$, $A_1$, $A_2$, and $\alpha$, we can explore the label hierarchy under the SEAL extension $(\mathcal{O}, \mathcal{L}, A_1, A_2, \alpha)$. This approach allows us to incorporate prior domain knowledge and guide the training of the model to improve performance while also providing a framework for interpreting the label hierarchy.

\paragraph{Random Walk Construction of $\alpha$.} \label{random walk}
We observe that the matrix $A$ satisfies the conditions in Theorem~\ref{thm:tree-assumption}, and can be viewed as a Markov transition matrix on $\mathcal{L}\cup\mathcal{O}$ that follows the top-down direction over a tree. Therefore, the probability of a random walk from a node $s\in \mathcal{L}$ to a node $r\in \mathcal{O}$ can be computed by simulating the Markov chain. We define $\alpha_{sr}$ to be the probability of random walks starting from $s$ and ending at $r$, which can be interpreted as the probability that node $r$ is contained in the subtree of node $s$. Specifically, we have:
\begin{align}
\alpha_{sr} = \left[ \sum_{k=1}^\infty A^k \right]_{sr} = \left[ \left(I-A\right)^{-1} \right]_{sr},
\end{align}
where $I$ is the identity matrix.

Moreover, we can further simplify the above equation by noting that $\left(I-A\right)^{-1}$ can be precomputed. Specifically, we have:
\begin{equation}
(I-A)^{-1}=
\begin{pmatrix}
(I-A_{1})^{-1} & (I-A_{1})^{-1}A_{2} \\
0 & I
\end{pmatrix}.
\end{equation}

This construction of $\alpha$ is based on the Markov chain on trees and has been applied in other applications such as document distance \cite{takezawa2021supervised} and hierarchical node clustering \cite{zugner2021end}. However, it is the first time that this construction has been used to train a deep neural network.

\paragraph{Optimizing SEAL regularization} Once $\alpha$ is defined explicitly through $A_2$, the expression of SEAL regularization $\Phi$ is also well defined. It simplifies to
\begin{align} \label{weighted l1 loss}
    &\phi(f_\theta(\bx),\delta_y) \nonumber\\
    = &(w^\top \left(\begin{array}{c}
\left(I-{A}_{1}\right)^{-1} {A}_{2} \\
I
\end{array}\right) \left(f_\theta(\bx) - \delta_y\right))^{|\cdot|},
\end{align}
where $^{|\cdot|}$ denotes taking the element-wise absolute value. We set $w = 1_N$ for simplicity.

One could jointly optimize $\theta$ and $A_2$ (or $\alpha$). Particular attention should be paid to $A_2$ since it is discrete and required to satisfy the conditions in Theorem~\ref{thm:tree-assumption}, making the optimization very hard. In this paper, we relax $A_2\in \{0, 1\}^{(N-K)\times K}$ to $A_2^{\rm soft} \in [0, 1]^{(N-K)\times K}$. One could employ projected gradient descent on each column of $A_2^{\rm soft}$ to ensure those conditions. More investigation on optimization could be found in Appendix~\ref{sec:update-rule}.

For clarity, we denote the SEAL regularization as $\phi_\Theta(f_\theta(x), \delta_y)$, where the suffix $\Theta$ denotes the parameters defining latent hierarchy.

\noindent\textbf{Interpreting SEAL results}
After $A_2^{\rm soft}$ is optimized, we can reconstruct $A_2$ to interpret the explored label hierarchy. Specifically
\begin{align}
    (A_{2})_{sr} = \left\{ \begin{array}{cc}
        1 & \text{if } s = \operatorname{argmax}_{k\in\mathcal{L}} (A^{soft}_{2})_{kr} \\
        0 & \textrm{otherwize}
    \end{array} \right.
\end{align}
Then the matrix $A$ is derived after optimization.

\subsection{Learning with SEAL}

We have already defined SEAL regularization based on the model output $f_\theta(\bx)$ and the target $y$. Then it is natural to apply SEAL regularization to various learning scenarios.

We consider the learning process in a typical mini-batch setting. Given a batch of samples $B\subset D$, we consider the averaged summation $\Phi$ of SEAL regularization $\phi$ over the batch.
\begin{align}
    \Phi(\theta, \Theta; B) = \frac{1}{|B|} \sum_{(\bx, y) \in B} \phi_\Theta(f_\theta(\bx), \delta_y).
\end{align}
We note that $\delta_y$ could be the one-hot labels of the labeled data or pseudo-labels on the unlabeled data.

\subsubsection{Supervised learning with SEAL}
Consider a supervised learning objective $L(\theta)$ over a batch, such as Cross-Entropy (CE) to train the neural network $\theta$. One could derive the SEAL regularized objective as

\begin{align}\label{eq:supervised-SEAL}
    \mathcal{L}(\theta, \Theta) & = \frac{1}{|B|} \sum_{(\bx, y)\in B} \operatorname{CE}(f_{\theta}(\bx),  \delta_y)) + \lambda \phi(f_{\theta}(\bx),  \delta_y))\nonumber\\
    & = L(\theta) + \lambda \Phi(\theta, \Theta; B)
\end{align}
Optimizing $\mathcal{L}(\theta, \Theta)$ jointly trains the neural network $f_\theta$ and the latent hierarchy defined by $\Theta$.

\subsubsection{Semi-supervised learning with SEAL}
Consider a general scheme~\cite{gong2021alphamatch} that unifies many prior semi-supervised algorithms. For the $n+1$-th iteration, the model parameter is derived based on supervised loss $L(\theta)$ and consistency regularization $\Psi(\theta; \Theta)$. Specifically
\begin{equation} 
\theta_{n+1} \leftarrow \underset{\theta}{\arg \min }\left\{L\left(\theta \right)+ \gamma \Psi(\theta; \theta_n) \right\},
\end{equation}
where $\theta_n$ denotes the model parameters at the $n$-th iteration and $\gamma$ is the loss balancing coefficient.

Adding SEAL to semi-supervised learning is no more than applying SEAL regularization to a supervised loss $L(\theta)$, which is shown in Eqn.~\eqref{eq:supervised-SEAL}, and consistency regularization $\Psi(\theta, \Theta)$, which will be described below.

Usually speaking, the computation $\Psi(\theta; \theta_n)$ is conducted over a batch of unlabeled data $B^u\subset D^u$. For each sample $\bx \in B^u$, the computation follows the following process:
\begin{compactdesc}
    \item[Pseudo-label prediction on weak augmentation] Computing the prediction of weakly-augmented image $\bx'\in P_{\rm wAug}(\bx)$ with model $f_{\theta_n}$ in the last iteration, which will be used to generate pseudo-labels in the next steps.
    \item[Strong augmentation] For the input data $\bx$, we sample strong argumentation $\bx''\in P_{\rm sAug}(x)$.
    \item[Selection] Some selection processes are applied to select the samples and assign them meaningful pseudo-labels $y'$s. This results in a new pseudo-labeled dataset $\hat B^u := \{ (\bx'', y') \}$.
\end{compactdesc}
Therefore, consistency regularization minimizes the differences between the prediction by the model and the pseudo-label $y'$, for example, using the cross entropy as follows:
\begin{align}
    \Psi(\theta, \theta_n) = \frac{1}{|\hat B^u|} \sum_{(\bx'', y')\in \hat B^u} \operatorname{CE}(f_{\theta_n}(\bx^{\prime \prime}), \delta_{y'}).
\end{align}

Similar to Eqn.~\eqref{eq:supervised-SEAL}, adding SEAL regularization is simply adding another term $\Phi(\theta, \Theta, \hat B^u)$. Then, we obtain the updating rule of semi-supervised learning with SEAL:
\begin{align}\label{eq:semi-supervised-SEAL}
    \theta_{n+1} \leftarrow \underset{\theta}{\arg \min }\Biggl\{& L\left(\theta \right)  + \gamma  \left[ \Psi(\theta; \theta_n) + \lambda \Phi(\theta, \Theta; \hat B^u)  \right]\Biggr\}.
\end{align}

Many existing approaches~\cite{sohn2020fixmatch,zhang2021flexmatch,wang2022debiased} fit into this paradigm. So SEAL can be applied easily to such approaches.

\noindent\textbf{Summary for SEAL}
We have demonstrated how to plugin SEAL with supervised and semi-supervised learning in Eqn.~(\ref{eq:supervised-SEAL}) and (\ref{eq:semi-supervised-SEAL}), respectively. The key observation is that SEAL regularization can be applied as long as there is an objective function between model prediction $f(\bx)$ and target $y$.

\section{Theoretical analysis of SEAL regularization}

We find Eqn.~\eqref{weighted l1 loss} has a similar structure to that of Tree-Wasserstein distance, so we shall first extend the definition of Tree-Wasserstein distance. 

We shall first rewrite the Tree-Wasserstein distance's summation using the node as indices. Note the lower endpoint of each edge has a one-to-one correspondence with each node, thus we can see the weight of each edge as the weight of each node. Denote the tree as $\mathcal{X}$ and leaf nodes as $\mathcal{X}_{leaf}$. We can rewrite the expression of Tree-Wasserstein distance into $W_{d_\mathcal{X}}(\mu, \nu)=\sum_{v \in \mathcal{X}} w_v|\mu(\Gamma(v))-\nu(\Gamma(v))|$. When $\mu$ and $\nu$ are supported only on the leaf set, we can rewrite $\Gamma(v)$ using the ancestor-child relationship. That is, 
\begin{equation} \label{hard-treew-subtreeprob}
\hspace{-2mm} W_{d_\mathcal{X}}(\mu, \nu) \hspace{-0.5mm} = \hspace{-0.5mm} \sum_{v \in \mathcal{X}} w_v| \hspace{-1mm} \sum_{x \in \mathcal{X}_{\text{leaf}}} \hspace{-1mm}(\mu(x) \hspace{-0.5mm}- \hspace{-0.5mm} \nu(x))\mathbb{I}_{\text{v is ancestor of x}}|.
\end{equation}

If a directed graph has its adjacency matrix $A$ satisfying the conditions in Theorem~\ref{thm:tree-assumption} except relaxing the hard constraint $\{0,1\}$ to $[0,1]$, we shall call it a soft tree. Recall the subtree probabilistic interpretation of $\alpha$ in \ref{random walk}, we can define relaxed Tree-Wasserstein distance (RTW) as below. 

\begin{definition}[Relaxed Tree-Wasserstein distance]
Assume $\mathcal{X}$ is a soft tree and denote the leaf nodes as $\mathcal{X}_{leaf}$. For any two probability measures supported on $\mathcal{X}_{\text{leaf}}$. The relaxed tree Wasserstein distance $W_{d_{\mathcal{X}}}^{\text{relax }}\left(\mu, \nu \right)$ is given as follows:
\begin{equation} \label{RTW definition}   
W_{d_{\mathcal{X}}}^{\text{relax }}\left(\mu, \nu \right) =\sum_{v \in \mathcal{X}} w_{v}\left|\sum_{x \in \mathcal{X}_{\text{leaf }}} \hspace{-1mm} \alpha_{vx}\left(\mu(x)-\nu(x)\right)\right|.
\end{equation}
\end{definition}

If we let $\mathcal{X}_{\text{leaf}}$ be the set of \emph{observed} labels $\mathcal{O}$ and $\mathcal{X}$ be \emph{total} labels $\mathcal{O} \cup \mathcal{L}$. We can then show the connection between the relaxed tree Wasserstein distance and the weighted total classification error given by Eqn.~(\ref{weighted l1 loss}).

\begin{theorem}
The weighted total classification loss described by Eqn.~(\ref{weighted l1 loss}) under $A^{soft}_2$ coincides with $W_{d_{\mathcal{X}}}^{\text {relax }}\left(f_{\theta}(\bx), \delta_{y} \right)$.
\end{theorem}

\begin{proof}
Please see Appendix \ref{Loss Equivalence RTW}.
\end{proof}

Here we would present some theoretical properties of relaxed Tree-Wasserstein distance next to illustrate why it is a good metric defined on trees.

\begin{theorem}
$W_{d_{\mathcal{X}}}^{\text {relax }}\left(\cdot, \cdot \right)$ defines a metric on the probability space. Furthermore, when $A$ is the (hard) adjacency matrix of a tree, the relaxed tree Wasserstein distance is exactly the tree Wasserstein distance.  
\end{theorem}

\begin{proof}
Please see Appendix \ref{Basic Property of RTW}.
\end{proof}

\begin{theorem}
The relaxed tree Wasserstein distance is a negative definite kernel.
\end{theorem}

\begin{proof}
Please see Appendix \ref{Kernel Property of RTW}.
\end{proof}

\section{Applying SEAL to semi-supervised learning} \label{sec:Experiments}

Firstly, SEAL improves standard supervised learning which outperforms label smoothing by a large margin, details could be found in Appendix \ref{sec:supervised}. Then we present our major experimental results on semi-supervised learning.

\subsection{Datasets}
We evaluate our proposed method on three popular datasets, namely CIFAR10, CIFAR100, and STL-10. 

\noindent\textbf{CIFAR10 and CIFAR100.}
CIFAR10\cite{krizhevsky2009learning} contains 60,000 colored images in 10 different classes, where each image has a size of $32 \times 32$ pixels. The training set consists of 50,000 labeled images and the test set consists of 10,000 labeled images. Similarly, CIFAR100\cite{krizhevsky2009learning} contains 100 different classes with the same image size and a similar number of images.

\noindent\textbf{STL-10.} STL-10\cite{coates2011analysis} is a semi-supervised benchmark that contains 10 classes. It is adapted from ImageNet\cite{deng2009imagenet} and contains 500 labeled training samples and 800 labeled testing samples per class. Additionally, it has 10,000 unlabeled images, some of which are not from the labeled classes.

Following the standard semi-supervised learning setting to sample, we sample the labeled images equally and randomly from all classes. To ensure statistical significance, we repeat each experiment five times and calculate the mean and standard deviation of the results.

\subsection{Baselines}
The baselines we consider in our experiments are those prior works similar to FixMatch, such as $\Pi$-Model \cite{laine2016temporal}, Pseudo Label \cite{lee2013pseudo}, Mean Teacher \cite{tarvainen2017mean}, MixMatch \cite{berthelot2019mixmatch}, ReMixMatch \cite{berthelot2019remixmatch}, VAT \cite{miyato2018virtual}, UDA \cite{xie2020unsupervised}, FlexMatch \cite{zhang2021flexmatch}and DebiasPL \cite{wang2022debiased}. However, we find that our proposed method SEAL is simple yet effective and outperforms all of these baselines on all three datasets in nearly all settings.

\subsection{Implementation of SEAL}
\noindent\textbf{Defining SEAL extension $(\mathcal{O}, \mathcal{L}, A_1, A_2, \alpha)$.} For fair comparison and injecting no prior knowledge, $A_1$ and $A^{soft}_2$ are both randomly initialized. For CIFAR10 and STL-10, $K=10$ and $N=21$. For CIFAR100, $K=100$ and $N=130$. More ablations and details can be found in Section \ref{initialize tree}.

\noindent\textbf{Combining SEAL with other learning methods.}
The default setting of SEAL for semi-supervised learning is adopted from the same configuration and hyper-parameters used in FixMatch\cite{sohn2020fixmatch}. SEAL (Curriculum) adopts the curriculum pseudo-labeling technique and hyper-parameters used in FlexMatch\cite{zhang2021flexmatch}. SEAL (Debiased) adopts the debiasing trick and hyper-parameters used in DebiasPL\cite{wang2022debiased}.

\noindent\textbf{Optimizing with SEAL regularization.}
Specifically, we use a (batch) stochastic gradient descent (SGD) optimizer with a momentum of 0.9. We set the learning rate scheduler as the cosine decay scheduler, where the learning rate $\beta$ can be expressed as $\beta=\beta_0 \mathbf{cos}(\frac{7\pi}{16}\frac{s}{S})$. Here, $\beta_0$ is the initial learning rate set to 0.03, $s$ is the current optimization step, and $S$ is the total number of optimization steps set to $2^{20}$. We set the batch size of the labeled training data to 64, and the ratio of unlabeled training data to labeled data $\mu$ is set to 7. We set the threshold $\tau$ to 0.95, and the weak and strong augmentation functions used in our experiments are based on RandAugment\cite{cubukpractical}. We use WideResNet-28-2 as the backbone model for our experiments.

\begin{table}[htb]
\centering
\begin{tabular}{@{}c|cc@{}}
\toprule
\multirow{2}{*}{Method}                       & \multicolumn{2}{c}{CIFAR10}                                             \\ \cmidrule(l){2-3} 
                                              & 40 labels                          & 250 labels                         \\ \midrule
$\Pi$-Model  & -                                  & $45.74 \pm 3.97$                   \\
ReMixMatch    & $80.90 \pm 9.64$                   & $94.56 \pm 0.05$                   \\
PseudoLabel          & -                                  & $50.22 \pm 0.43$                   \\
MeanTeacher     & -                                  & $67.68 \pm 2.30$                   \\
MixMatch         & $52.46 \pm 11.5$                   & $88.95 \pm 0.86$                   \\
VAT              & $25.34 \pm 2.12$                   & $58.97 \pm 1.79$                   \\
UDA               & $70.95 \pm 5.93$                   & $91.18 \pm 1.08$                   \\
FixMatch             & $86.19 \pm 3.37$                   & $94.93 \pm 0.65$                   \\
FlexMatch            & $95.03 \pm 0.06$                   & $95.02 \pm 0.09$                   \\
DebiasPL               & $94.60 \pm 1.30$                    & $95.40 \pm 0.10$                    \\ \midrule
\textbf{SEAL}                     & $\mathbf{93.71} \pm \mathbf{0.58}$ & $\mathbf{95.57} \pm \mathbf{0.55}$ \\
\textbf{SEAL (Debiased)}                     & $\mathbf{95.34} \pm \mathbf{0.07}$ & $\mathbf{95.59} \pm \mathbf{0.18}$ \\ \midrule
\end{tabular}
\caption{Results on CIFAR10 dataset}
\label{tab:results-cifar10}
\end{table}

\begin{table}[htb]
\centering
\begin{tabular}{@{}c|cc@{}}
\toprule
\multirow{2}{*}{Method}                       & \multicolumn{2}{c}{CIFAR100}                                             \\ \cmidrule(l){2-3} 
                                              & 400 labels                          & 2500 labels                         \\ \midrule
$\Pi$-Model & -                                  & $42.75 \pm 0.48$                   \\
ReMixMatch     & $55.72 \pm 2.06$                   & $72.57 \pm 0.31$                   \\
PseudoLabel            & -                                  & $42.62 \pm 0.46$                   \\
MeanTeacher         & -                                  & $46.09 \pm 0.57$                   \\
MixMatch        & $32.39 \pm 1.32$                   & $60.06 \pm 0.37$                   \\
VAT            & $51.15 \pm 1.75$                   & $53.16 \pm 0.79$                   \\
UDA           & $40.72 \pm 0.88$                   & $66.87 \pm 0.22$                   \\
FixMatch       & $51.15 \pm 1.75$                   & $71.71 \pm 0.11$                   \\
FlexMatch      & $60.06 \pm 1.62$                   & $73.51 \pm 0.20$                   \\
 \midrule
\textbf{SEAL}                     & $\mathbf{54.82} \pm \mathbf{0.79}$ & $\mathbf{73.01} \pm \mathbf{0.23}$ \\
\textbf{SEAL (Curriculum)}                     & $\mathbf{60.27} \pm \mathbf{1.58}$ & $\mathbf{73.61} \pm \mathbf{0.29}$ \\ \midrule
\end{tabular}
\caption{Results on CIFAR100 dataset}
\label{tab:results-cifar100}
\end{table}

\begin{table}[htb]
\centering
\begin{tabular}{@{}c|cc@{}}
\toprule
\multirow{2}{*}{Method}                       & \multicolumn{2}{c}{STL-10}                                             \\ \cmidrule(l){2-3} 
                                              & 40 labels                          & 250 labels                         \\ \midrule
$\Pi$-Model & $25.69 \pm 0.85$                                  & $44.87 \pm 1.50$                   \\
ReMixMatch      & $67.88 \pm 6.24$                   & $87.51 \pm 1.28$                   \\
PseudoLabel               & $25.32 \pm 0.99$                                  & $44.55 \pm 2.43$                   \\
MeanTeacher           & $28.28 \pm 1.45$                                  & $43.51 \pm 2.75$                   \\
MixMatch          & $45.07 \pm 0.96$                   & $65.48 \pm 0.32$                   \\
VAT                   & $25.26 \pm 0.38$                   & $43.58 \pm 1.97$                   \\
UDA                 & $62.58 \pm 8.44$                   & $90.28 \pm 1.15$                   \\
FixMatch               & $64.03 \pm 4.14$                   & $90.19 \pm 1.04$                   \\
FlexMatch               & $70.85 \pm 4.16$                   & $91.77 \pm 0.39$                   \\
 \midrule
\textbf{SEAL}                     & $\mathbf{66.73} \pm \mathbf{3.21}$ & $\mathbf{90.23} \pm \mathbf{0.54}$ \\ 
\textbf{SEAL (Curriculum)}                     & $\mathbf{83.85} \pm \mathbf{2.63}$ & $\mathbf{92.08} \pm \mathbf{0.47}$ \\ \midrule
\end{tabular}
\caption{Results on STL-10 dataset}
\label{tab:results-stl-10}
\end{table}

\subsection{Findings}

\noindent\textbf{SEAL is simple yet effective.} SEAL is easy to implement and with the aid of SEAL and its variants, we can achieve state-of-art results on all three datasets under all label amount settings.

\noindent\textbf{The fewer labeled data, the more significant improvements.} Interestingly, we observe that the fewer labeled data available, the more significant gains we can achieve using SEAL. For instance, on CIFAR10, we obtain a remarkable $7.52$ accuracy gain with only $40$ labeled data, while we only see a $0.64$ accuracy gain with $250$ labeled data. This finding highlights the effectiveness of our proposed method in situations where labeled data is scarce.

\noindent\textbf{SEAL can be boosted by various techniques.} Moreover, we demonstrate that our proposed method can be further enhanced by incorporating various existing semi-supervised learning techniques, such as Curriculum Pseudo Label~\cite{zhang2021flexmatch} and Debiased Pseudo Label~\cite{wang2022debiased}, into SEAL framework with minimal effort. This implies that any future work on improving the quality of pseudo labels can be easily adapted into our SEAL framework.

\subsection{Ablation Studies \label{sec:Ablation}}

In this section, we focus on analyzing the influence of different parameters on the performance of our proposed method, SEAL (Debiased), using the CIFAR10 dataset with only $40$ labeled samples.

\subsubsection{Different tree structure} \label{initialize tree}
\begin{table}[ht]
\vskip-.5cm

\centering
\begin{tabular}{l|l}
\toprule
Tree Name & Acc. on CIFAR10 (40 labels)   \\
\midrule
Without Tree & 94.60    \\
Trivial Tree & 95.18    \\
Random Tree & 95.34   \\
NBDT \cite{wan2020nbdt} Tree & 95.39   \\
\bottomrule
\end{tabular}
\caption{\label{tree-structure} Classification under different tree structure}
\vskip-.5cm
\end{table}

Next, we examine how the choice of tree structure affects the results. We compare three different trees: a trivial tree with all $10$ classes as leaf nodes besides one root node, a randomly generated depth-$4$ tree with $21$ nodes, and the NBDT tree proposed in \cite{wan2020nbdt}, which has a well-designed hierarchy as shown in Figure \ref{nbdt-tree}. We use the adjacency matrix of the internal nodes induced subtree as the adjacency matrix $A_1$ in our method for each tree.

The results of the classification accuracies under these different trees are presented in Table \ref{tree-structure}. As we can see from the table, using tree structures consistently improves the classification accuracies compared to the vanilla cases. It is worth noting that the NBDT tree, which is carefully designed, achieves the highest accuracy, while our randomly generated tree performs better than the other cases.

\subsubsection{Different regularizer}
\begin{figure}[H] 
\centering 
\includegraphics[width=0.6\columnwidth]{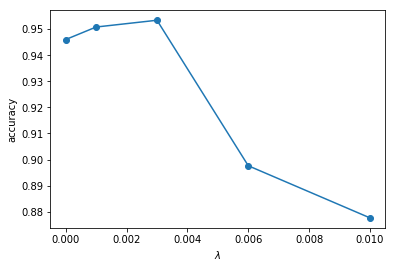} 
\caption{The influence of different hyper-parameter} 
\label{regular} 
\end{figure}

Lastly, we explore the impact of the regularizer on classification accuracy by experimenting with $5$ different $\lambda$ values, including the case of $\lambda=0$ to illustrate the necessity of the regularizer. As shown in Figure~\ref{regular}, when $\lambda$ approaches $0.003$, the accuracy increases, while it decreases as $\lambda$ deviates from $0.003$. Therefore, we conclude that the optimal value of $\lambda$ is around $0.003$. We note that the accuracy drops significantly when $\lambda$ is too large, which may be attributed to the imbalance in loss scale.

\section{Conclusion and Future Work}

In this paper, we propose a framework SEAL to jointly train the model of high performances and the label structure of significance. The SEAL framework is flexible to be adapted to various learning schemes, and can even incorporate the prior structure given by the external knowledge and the information given by the data. Experimental results support the effectiveness of the SEAL framework. Theoretical understanding of SEAL via optimal transport theory is also discussed. Future works may include incorporating more complex prior knowledge or applying the SEAL framework to self-supervised learning.

{\small
\bibliographystyle{ieee_fullname}
\bibliography{egbib}

\begin{thebibliography}{10}\itemsep=-1pt

\bibitem{assran2021semi}
Mahmoud Assran, Mathilde Caron, Ishan Misra, Piotr Bojanowski, Armand Joulin,
  Nicolas Ballas, and Michael Rabbat.
\newblock Semi-supervised learning of visual features by non-parametrically
  predicting view assignments with support samples.
\newblock In {\em Proceedings of the IEEE/CVF International Conference on
  Computer Vision}, pages 8443--8452, 2021.

\bibitem{barz2019hierarchy}
Bj{\"o}rn Barz and Joachim Denzler.
\newblock Hierarchy-based image embeddings for semantic image retrieval.
\newblock In {\em 2019 IEEE Winter Conference on Applications of Computer
  Vision (WACV)}, pages 638--647. IEEE, 2019.

\bibitem{bengio2010label}
Samy Bengio, Jason Weston, and David Grangier.
\newblock Label embedding trees for large multi-class tasks.
\newblock {\em Advances in Neural Information Processing Systems}, 23, 2010.

\bibitem{berg1984harmonic}
Christian Berg, Jens Peter~Reus Christensen, and Paul Ressel.
\newblock {\em Harmonic analysis on semigroups: theory of positive definite and
  related functions}, volume 100.
\newblock Springer, 1984.

\bibitem{berthelot2019remixmatch}
David Berthelot, Nicholas Carlini, Ekin~D Cubuk, Alex Kurakin, Kihyuk Sohn, Han
  Zhang, and Colin Raffel.
\newblock Remixmatch: Semi-supervised learning with distribution alignment and
  augmentation anchoring.
\newblock {\em arXiv preprint arXiv:1911.09785}, 2019.

\bibitem{berthelot2019mixmatch}
David Berthelot, Nicholas Carlini, Ian Goodfellow, Nicolas Papernot, Avital
  Oliver, and Colin~A Raffel.
\newblock Mixmatch: A holistic approach to semi-supervised learning.
\newblock {\em Advances in neural information processing systems}, 32, 2019.

\bibitem{bertinetto2020making}
Luca Bertinetto, Romain Mueller, Konstantinos Tertikas, Sina Samangooei, and
  Nicholas~A Lord.
\newblock Making better mistakes: Leveraging class hierarchies with deep
  networks.
\newblock In {\em Proceedings of the IEEE/CVF Conference on Computer Vision and
  Pattern Recognition}, pages 12506--12515, 2020.

\bibitem{bilal2017convolutional}
Alsallakh Bilal, Amin Jourabloo, Mao Ye, Xiaoming Liu, and Liu Ren.
\newblock Do convolutional neural networks learn class hierarchy?
\newblock {\em IEEE transactions on visualization and computer graphics},
  24(1):152--162, 2017.

\bibitem{chang2021your}
Dongliang Chang, Kaiyue Pang, Yixiao Zheng, Zhanyu Ma, Yi-Zhe Song, and Jun
  Guo.
\newblock Your" flamingo" is my" bird": Fine-grained, or not.
\newblock In {\em Proceedings of the IEEE/CVF Conference on Computer Vision and
  Pattern Recognition}, pages 11476--11485, 2021.

\bibitem{chen2020big}
Ting Chen, Simon Kornblith, Kevin Swersky, Mohammad Norouzi, and Geoffrey~E
  Hinton.
\newblock Big self-supervised models are strong semi-supervised learners.
\newblock {\em Advances in neural information processing systems},
  33:22243--22255, 2020.

\bibitem{coates2011analysis}
Adam Coates, Andrew Ng, and Honglak Lee.
\newblock An analysis of single-layer networks in unsupervised feature
  learning.
\newblock In {\em Proceedings of the fourteenth international conference on
  artificial intelligence and statistics}, pages 215--223. JMLR Workshop and
  Conference Proceedings, 2011.

\bibitem{cubukpractical}
Ekin~Dogus Cubuk, Barret Zoph, Jon Shlens, and Randaugment Le~QV.
\newblock Practical automated data augmentation with a reduced search space.
\newblock In {\em Proceedings of the IEEE/CVF Conference on Computer Vision and
  Pattern Recognition Workshops}, pages 702--703.

\bibitem{deng2010does}
Jia Deng, Alexander~C Berg, Kai Li, and Li Fei-Fei.
\newblock What does classifying more than 10,000 image categories tell us?
\newblock In {\em European conference on computer vision}, pages 71--84.
  Springer, 2010.

\bibitem{deng2009imagenet}
Jia Deng, Wei Dong, Richard Socher, Li-Jia Li, Kai Li, and Li Fei-Fei.
\newblock Imagenet: A large-scale hierarchical image database.
\newblock In {\em 2009 IEEE conference on computer vision and pattern
  recognition}, pages 248--255. Ieee, 2009.

\bibitem{deng2012hedging}
Jia Deng, Jonathan Krause, Alexander~C Berg, and Li Fei-Fei.
\newblock Hedging your bets: Optimizing accuracy-specificity trade-offs in
  large scale visual recognition.
\newblock In {\em 2012 IEEE Conference on Computer Vision and Pattern
  Recognition}, pages 3450--3457. IEEE, 2012.

\bibitem{frome2013devise}
Andrea Frome, Greg~S Corrado, Jon Shlens, Samy Bengio, Jeff Dean, Marc'Aurelio
  Ranzato, and Tomas Mikolov.
\newblock Devise: A deep visual-semantic embedding model.
\newblock {\em Advances in neural information processing systems}, 26, 2013.

\bibitem{garg2022hiermatch}
Ashima Garg, Shaurya Bagga, Yashvardhan Singh, and Saket Anand.
\newblock Hiermatch: Leveraging label hierarchies for improving semi-supervised
  learning.
\newblock In {\em Proceedings of the IEEE/CVF Winter Conference on Applications
  of Computer Vision}, pages 1015--1024, 2022.

\bibitem{garg2022learning}
Ashima Garg, Depanshu Sani, and Saket Anand.
\newblock Learning hierarchy aware features for reducing mistake severity.
\newblock {\em arXiv preprint arXiv:2207.12646}, 2022.

\bibitem{garnot2020leveraging}
Vivien Sainte~Fare Garnot and Loic Landrieu.
\newblock Leveraging class hierarchies with metric-guided prototype learning.
\newblock {\em arXiv preprint arXiv:2007.03047}, 2020.

\bibitem{gong2021alphamatch}
Chengyue Gong, Dilin Wang, and Qiang Liu.
\newblock Alphamatch: Improving consistency for semi-supervised learning with
  alpha-divergence.
\newblock In {\em Proceedings of the IEEE/CVF Conference on Computer Vision and
  Pattern Recognition}, pages 13683--13692, 2021.

\bibitem{hu2021simple}
Zijian Hu, Zhengyu Yang, Xuefeng Hu, and Ram Nevatia.
\newblock Simple: similar pseudo label exploitation for semi-supervised
  classification.
\newblock In {\em Proceedings of the IEEE/CVF Conference on Computer Vision and
  Pattern Recognition}, pages 15099--15108, 2021.

\bibitem{jeevan2021vision}
Pranav Jeevan and Amit Sethi.
\newblock Vision xformers: Efficient attention for image classification.
\newblock {\em arXiv preprint arXiv:2107.02239}, 2021.

\bibitem{karthik2021no}
Shyamgopal Karthik, Ameya Prabhu, Puneet~K Dokania, and Vineet Gandhi.
\newblock No cost likelihood manipulation at test time for making better
  mistakes in deep networks.
\newblock {\em arXiv preprint arXiv:2104.00795}, 2021.

\bibitem{kontschieder2015deep}
Peter Kontschieder, Madalina Fiterau, Antonio Criminisi, and Samuel~Rota Bulo.
\newblock Deep neural decision forests.
\newblock In {\em Proceedings of the IEEE international conference on computer
  vision}, pages 1467--1475, 2015.

\bibitem{krizhevsky2009learning}
Alex Krizhevsky, Geoffrey Hinton, et~al.
\newblock Learning multiple layers of features from tiny images.
\newblock {\em .}, 2009.

\bibitem{laine2016temporal}
Samuli Laine and Timo Aila.
\newblock Temporal ensembling for semi-supervised learning.
\newblock {\em arXiv preprint arXiv:1610.02242}, 2016.

\bibitem{le2019tree}
Tam Le, Makoto Yamada, Kenji Fukumizu, and Marco Cuturi.
\newblock Tree-sliced variants of wasserstein distances.
\newblock {\em Advances in neural information processing systems}, 32, 2019.

\bibitem{lee2013pseudo}
Dong-Hyun Lee et~al.
\newblock Pseudo-label: The simple and efficient semi-supervised learning
  method for deep neural networks.
\newblock In {\em Workshop on challenges in representation learning, ICML},
  volume~3, page 896, 2013.

\bibitem{li2022maxmatch}
Yangbangyan Jiang~Xiaodan Li, Yuefeng Chen, Yuan He, Qianqian Xu, Zhiyong Yang,
  Xiaochun Cao, and Qingming Huang.
\newblock Maxmatch: Semi-supervised learning with worst-case consistency.
\newblock {\em IEEE Transactions on Pattern Analysis and Machine Intelligence},
  2022.

\bibitem{liu2020hyperbolic}
Shaoteng Liu, Jingjing Chen, Liangming Pan, Chong-Wah Ngo, Tat-Seng Chua, and
  Yu-Gang Jiang.
\newblock Hyperbolic visual embedding learning for zero-shot recognition.
\newblock In {\em Proceedings of the IEEE/CVF conference on computer vision and
  pattern recognition}, pages 9273--9281, 2020.

\bibitem{miller1998wordnet}
George~A Miller.
\newblock {\em WordNet: An electronic lexical database}.
\newblock MIT press, 1998.

\bibitem{miyato2018virtual}
Takeru Miyato, Shin-ichi Maeda, Masanori Koyama, and Shin Ishii.
\newblock Virtual adversarial training: a regularization method for supervised
  and semi-supervised learning.
\newblock {\em IEEE transactions on pattern analysis and machine intelligence},
  41(8):1979--1993, 2018.

\bibitem{nassar2021all}
Islam Nassar, Samitha Herath, Ehsan Abbasnejad, Wray Buntine, and Gholamreza
  Haffari.
\newblock All labels are not created equal: Enhancing semi-supervision via
  label grouping and co-training.
\newblock In {\em Proceedings of the IEEE/CVF Conference on Computer Vision and
  Pattern Recognition}, pages 7241--7250, 2021.

\bibitem{peyre2019computational}
Gabriel Peyr{\'e}, Marco Cuturi, et~al.
\newblock Computational optimal transport: With applications to data science.
\newblock {\em Foundations and Trends{\textregistered} in Machine Learning},
  11(5-6):355--607, 2019.

\bibitem{Rasmus2015SemiSupervisedLW}
Antti Rasmus, Harri Valpola, Mikko Honkala, Mathias Berglund, and Tapani Raiko.
\newblock Semi-supervised learning with ladder network.
\newblock {\em ArXiv}, abs/1507.02672, 2015.

\bibitem{sohn2020fixmatch}
Kihyuk Sohn, David Berthelot, Nicholas Carlini, Zizhao Zhang, Han Zhang,
  Colin~A Raffel, Ekin~Dogus Cubuk, Alexey Kurakin, and Chun-Liang Li.
\newblock Fixmatch: Simplifying semi-supervised learning with consistency and
  confidence.
\newblock {\em Advances in neural information processing systems}, 33:596--608,
  2020.

\bibitem{speer2017conceptnet}
Robyn Speer, Joshua Chin, and Catherine Havasi.
\newblock Conceptnet 5.5: An open multilingual graph of general knowledge.
\newblock In {\em Thirty-first AAAI conference on artificial intelligence},
  2017.

\bibitem{takezawa2021supervised}
Yuki Takezawa, Ryoma Sato, and Makoto Yamada.
\newblock Supervised tree-wasserstein distance.
\newblock In {\em International Conference on Machine Learning}, pages
  10086--10095. PMLR, 2021.

\bibitem{tarvainen2017mean}
Antti Tarvainen and Harri Valpola.
\newblock Mean teachers are better role models: Weight-averaged consistency
  targets improve semi-supervised deep learning results.
\newblock {\em Advances in neural information processing systems}, 30, 2017.

\bibitem{wan2020nbdt}
Alvin Wan, Lisa Dunlap, Daniel Ho, Jihan Yin, Scott Lee, Henry Jin, Suzanne
  Petryk, Sarah~Adel Bargal, and Joseph~E Gonzalez.
\newblock Nbdt: neural-backed decision trees.
\newblock {\em arXiv preprint arXiv:2004.00221}, 2020.

\bibitem{wang2021data}
Xudong Wang, Long Lian, and Stella~X Yu.
\newblock Data-centric semi-supervised learning.
\newblock {\em arXiv preprint arXiv:2110.03006}, 2021.

\bibitem{wang2022debiased}
Xudong Wang, Zhirong Wu, Long Lian, and Stella~X Yu.
\newblock Debiased learning from naturally imbalanced pseudo-labels.
\newblock In {\em Proceedings of the IEEE/CVF Conference on Computer Vision and
  Pattern Recognition}, pages 14647--14657, 2022.

\bibitem{wang2021hierarchical}
Yu Wang, Zhou Wang, Qinghua Hu, Yucan Zhou, and Honglei Su.
\newblock Hierarchical semantic risk minimization for large-scale
  classification.
\newblock {\em IEEE Transactions on Cybernetics}, 2021.

\bibitem{wu2016learning}
Hui Wu, Michele Merler, Rosario Uceda-Sosa, and John~R Smith.
\newblock Learning to make better mistakes: Semantics-aware visual food
  recognition.
\newblock In {\em Proceedings of the 24th ACM international conference on
  Multimedia}, pages 172--176, 2016.

\bibitem{xie2020unsupervised}
Qizhe Xie, Zihang Dai, Eduard Hovy, Thang Luong, and Quoc Le.
\newblock Unsupervised data augmentation for consistency training.
\newblock {\em Advances in Neural Information Processing Systems},
  33:6256--6268, 2020.

\bibitem{xu2021dash}
Yi Xu, Lei Shang, Jinxing Ye, Qi Qian, Yu-Feng Li, Baigui Sun, Hao Li, and Rong
  Jin.
\newblock Dash: Semi-supervised learning with dynamic thresholding.
\newblock In {\em International Conference on Machine Learning}, pages
  11525--11536. PMLR, 2021.

\bibitem{zhang2021flexmatch}
Bowen Zhang, Yidong Wang, Wenxin Hou, Hao Wu, Jindong Wang, Manabu Okumura, and
  Takahiro Shinozaki.
\newblock Flexmatch: Boosting semi-supervised learning with curriculum pseudo
  labeling.
\newblock {\em Advances in Neural Information Processing Systems},
  34:18408--18419, 2021.

\bibitem{zugner2021end}
Daniel Z{\"u}gner, Bertrand Charpentier, Morgane Ayle, Sascha Geringer, and
  Stephan G{\"u}nnemann.
\newblock End-to-end learning of probabilistic hierarchies on graphs.
\newblock In {\em International Conference on Learning Representations}, 2021.

\end{thebibliography}
}

\clearpage
\newpage
\appendix

\onecolumn

\section{More Ablation Studies}

\subsection{Different threshold}
\begin{figure}[H] 
\centering 
\includegraphics[width=0.4\columnwidth]{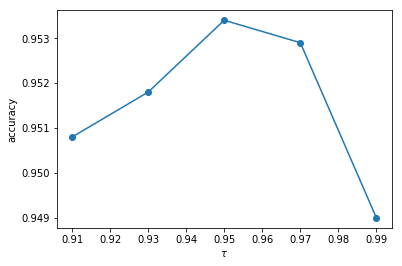} 
\caption{Varying different threshold for pseudo labeling} 
\label{thresh} 
\end{figure}

We first investigate the impact of threshold choice on the accuracy of our method. We experiment with five different values of $\tau$ and plot the results in Figure \ref{thresh}. As shown in the figure, the accuracy increases as $\tau$ approach $0.95$, and decreases when $\tau$ deviates from $0.95$. This suggests that choosing a threshold of $0.95$ yields the best performance.

\subsection{Convergence Speedup}

\begin{figure}[htb]
\centering
\includegraphics[width=0.4\columnwidth]{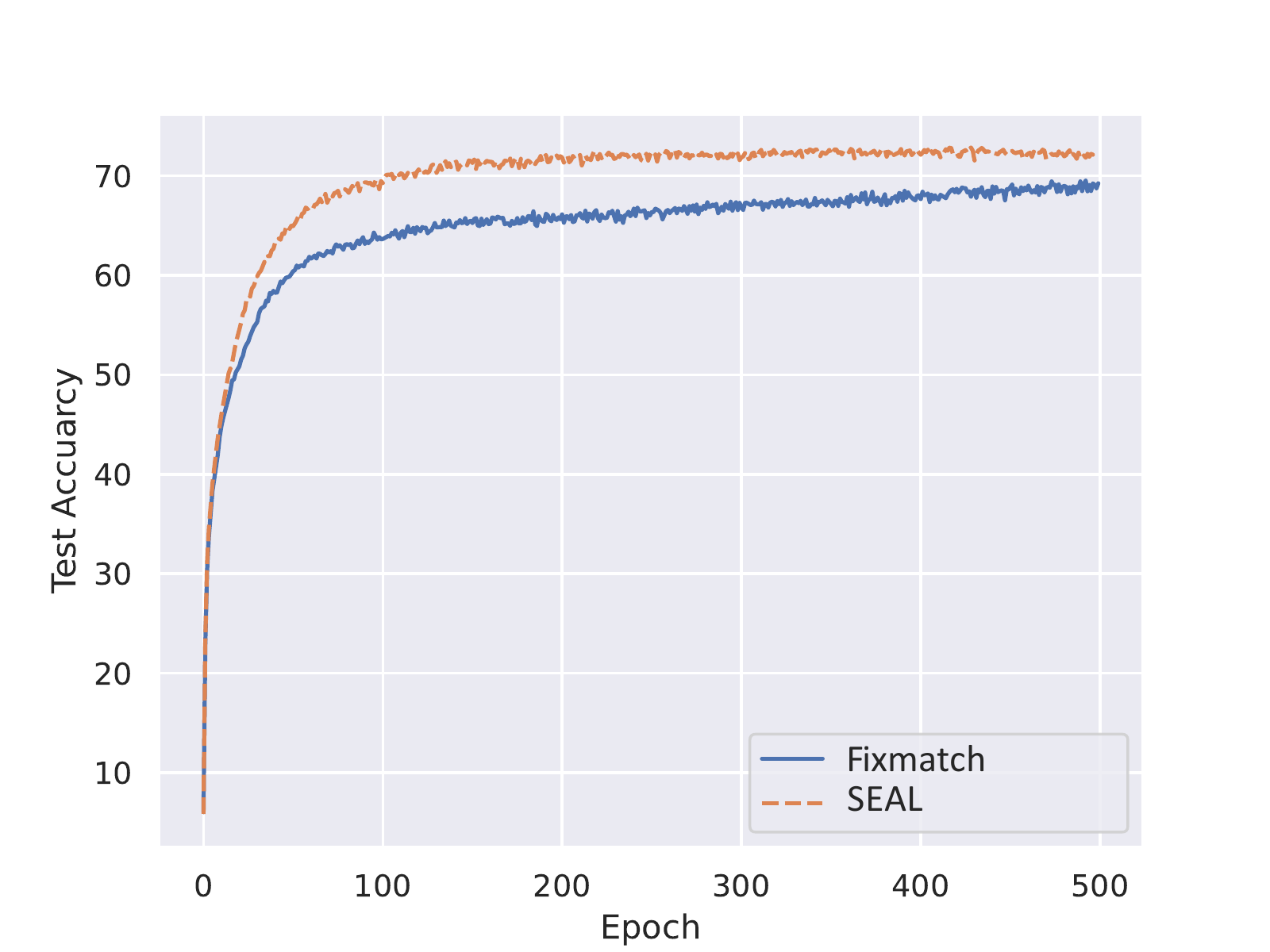}
\caption{Top-1 accuracy on CIFAR100 dataset (2500 labels)}
\label{convergence}
\end{figure}

Figure \ref{convergence} displays the top-1 accuracy of CIFAR100-2500 labels, showcasing that with the aid of SEAL, convergence becomes faster and more stable. With the addition of SEAL, the model consistently outperforms the initial training process in all epochs.

\subsection{Computation overhead of introducing SEAL}

We investigate the additional computation time required when applying SEAL to our method. Table \ref{computation-time} shows the results of these experiments, which were conducted using an Nvidia GeForce RTX 2080 Ti. We observe that the computation of SEAL incurs only a marginal increase in the computation time, demonstrating its efficiency in practice.

\begin{table}[htb]
\caption{\label{computation-time} Computation Time of $1$ epoch}
\centering
\begin{tabular}{ll}
\toprule
Update Method & Computation Time (Minutes)    \\
\midrule
with SEAL & 7.11   \\
without SEAL & 7.10  \\
\bottomrule
\end{tabular}
\end{table}

\subsection{Different updating rule for the adjacency matrix}\label{sec:update-rule}
Efficient updating of the soft adjacency matrix $A_2$ is essential in the experiments. Two popular approaches have been used for updating $A_2$. One approach is to consider each column of $A_2$ as the realization of softmax mapping. The other approach is to use projected gradient descent (PGD) to update the matrix, projecting each column onto the probability simplex. The classification accuracies of both methods are summarized in Table \ref{update-method}.

\begin{table}[htb]
\caption{\label{update-method} Different methods for updating adjacency matrix}
\centering
\begin{tabular}{ll}
\toprule
Update Method & CIFAR10-40 labels    \\
\midrule
PGD & 95.34   \\
Softmax & 86.46   \\
\bottomrule
\end{tabular}
\end{table}

It is evident that PGD has a significant advantage over the softmax mapping-based approach. This observation is also reported in the paper \cite{zugner2021end}. The poor performance of the softmax-based approach may be attributed to a bad initialization, where the optimization is trapped by the bad starting point.

\begin{figure}[htb] 
\centering 
\includegraphics[width=0.4\textwidth]{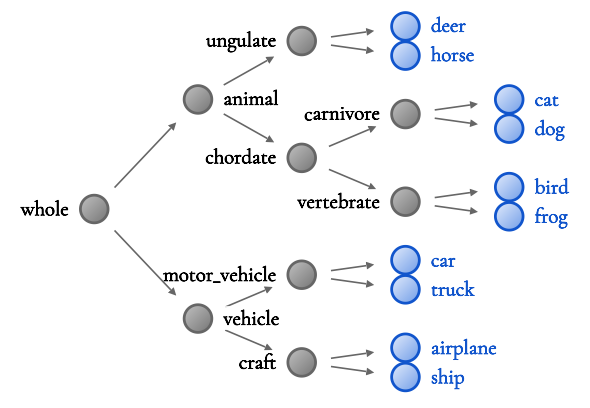} 
\caption{The hierarchy of NBDT Tree} 
\label{nbdt-tree} 
\end{figure}

Fig.\ref{nbdt-tree} shows the hierarchy of the NBDT tree.


\section{More Theoretical Results}

\subsection{Optimal Transport}

\begin{definition}[Wasserstein-$1$ distance] \label{wass def}
Consider two probability distribution: $\boldsymbol{x} \sim \mu$, and $\boldsymbol{y} \sim \nu$. The Wasserstein-$1$ distance between $\mu$ and $\nu$ can be defined as:
$$
\mathcal{W}\left(\mu, \nu\right)=\min _{\pi \in \Pi\left(\mu, \nu\right)} \int_{\mathcal{X} \times \mathcal{X}} c\left(\boldsymbol{x}, \boldsymbol{y}\right) d \pi,
$$
where $\mathcal{X}$ is the space that $\mu$ and $\nu$ supported on, $c(\cdot, \cdot)$ is a cost function defined on the cartesian space $\mathcal{X} \times \mathcal{X}$, and $\Pi\left(\mu, \nu\right)$ is the set of all possible couplings of $\mu$ and $\nu$ ; and $\pi$ is a joint distribution satisfying $\int_{\mathcal{X}} \pi\left(\boldsymbol{x}, \boldsymbol{y}\right) d \boldsymbol{y}=\mu\left(\boldsymbol{x}\right)$ and $\int_{\mathcal{X}} \pi\left(\boldsymbol{x}, \boldsymbol{y}\right) d \boldsymbol{x}= \nu\left(\boldsymbol{y}\right)$.
\end{definition}

\subsection{The Relation between SEAL and RTW \label{Loss Equivalence RTW}}

Note that we assume no correlation between real labels, thus the vector constructed by $\Pr(r|{y})$ as its $r$-th component is exactly $\delta_y$.

As the nodes in $\mathcal{O} \cup \mathcal{L}$ are exactly all the \emph{total} labels, and those in $\mathcal{O}$ are all the \emph{observed} labels. Comparing Eqn.~(\ref{weighted l1 loss}) and (\ref{RTW definition}) yields the result. For convenience, we consider the entire tree metric space as $\mathcal{X}$. Then it is easy to see that $\mathcal{X} = \mathcal{O} \cup \mathcal{L}$, and $\mathcal{X}_{\rm leaf} = \mathcal{O}$ are the leaf nodes of the tree.

\subsection{Basic Property of RTW \label{Basic Property of RTW}}

The positive definiteness and symmetry of $W_{d_{\mathcal{X}}}^{\text {relax }}$ is clear from it's definition. Then we show it satisfies the triangular inequality. For any probability measures $\lambda$, $\mu$ and $\nu$ on $\mathcal{X}_{\text{leaf}}$.
\begin{align}
&W_{d_{\mathcal{X}}}^{\text {relax }}\left(\lambda, \mu \right)+ W_{d_{\mathcal{X}}}^{\text {relax }}\left(\mu, \nu \right) \nonumber \\
&= \sum_{v \in \mathcal{X}} w_{v}\left|\sum_{x \in \mathcal{X}_{\text {leaf }}} \alpha_{vx} \left(\lambda(x)-\mu(x)\right)\right|  \nonumber \\
&+ \sum_{v \in \mathcal{X}} w_{v}\left|\sum_{x \in \mathcal{X}_{\text {leaf }}} \alpha_{vx}\left(\mu(x)-\nu(x)\right)\right|  \nonumber \\
& \geq \sum_{v \in \mathcal{X}} w_{v}\left|\sum_{x \in \mathcal{X}_{\text {leaf }}} \alpha_{vx}\left(\lambda(x)-\nu(x)\right)\right| \nonumber \\
& = W_{d_{\mathcal{X}}}^{\text {relax }}\left(\lambda, \nu \right).
\end{align}

Note that when $\mathcal{X}$ is a hard tree, we have $\mathbb{I}_{\text{v is the ancestor of x}}=\alpha_{vx}$. Then from equations (\ref{hard-treew-subtreeprob}) and (\ref{RTW definition}), we know that RTW is exactly the tree Wasserstein distance in this degenerate case.

\subsection{Kernel Property of RTW \label{Kernel Property of RTW}}

\begin{definition}\cite{berg1984harmonic}
A function $k: \mathcal{M} \times \mathcal{M} \rightarrow \mathbb{R}$ is negative definite if for $\forall n \geq 2$, $\forall x_1, x_2, \ldots, x_n \in \mathcal{M}$ and $\forall c_i \in \mathbb{R}$ such that $\sum^{n}_{i=1} c_i=0$, we have $\sum_{i, j} c_i c_j k\left(x_i, x_j\right) \leq 0$.
\end{definition}

We shall prove that RTW is a negative definite kernel on the tree leaf Wasserstein space $\mathcal{M}=\mathbb{P}(\mathcal{X}_{\text{leaf}})$. We define a mapping $\Phi$ where
\begin{equation}
\Phi(x)= 
(\begin{array}{c}
\left(I-{A}_{1}\right)^{-1} {A}_{2} \\
I
\end{array})x.    
\end{equation}

Note $k(x_i,x_j)= \left\| w \circ \Phi(x_i) - w \circ \Phi(x_j) \right\|_1$. Since the definition of negative definiteness is only related to the value of $k$, thus we can transform the problem of considering $\Phi$ only. Note $k$ is only a weighted $l_1$ distance between $\Phi$, from  the separability of $l_1$ norm and \cite{le2019tree}'s Lemma A.2, it is clear that RTW is negative definite.

\section{A Closer Look at the Supervised Settings}
\label{sec:supervised}
In this section, we shall show the performance of SEAL regularization on two backbones. One is a backbone with fewer parameters, another is the standard ResNet 18 backbone. 
\subsection{ViN backbone}
We train Vision Nystromformer
(ViN) \cite{jeevan2021vision} with optimizer AdamW for $60$ epochs and get classification accuracy $66.65\%$, apply the same configuration to label smoothing will give an accuracy of $68.02\%$. SEAL regularization boosts the accuracy of initial ViN from $66.65\%$ to $68.52\%$ within $5$ epochs.

\begin{figure}[htb]
\centering
\begin{subfigure}[b]{0.475\columnwidth}
\includegraphics[width=\textwidth]{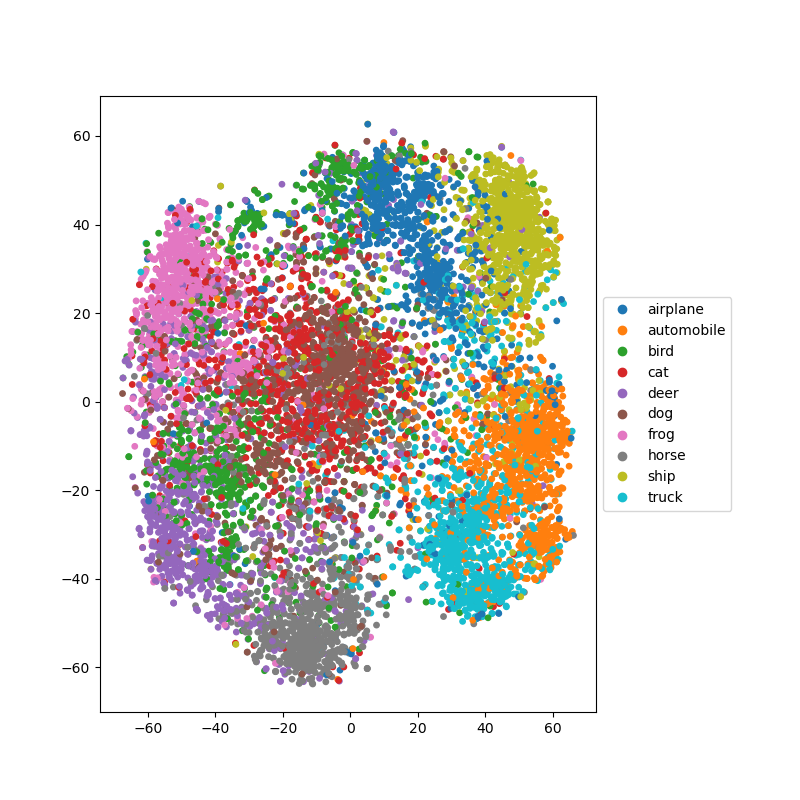}
\caption{t-SNE visualization of the initial feature of ViN backbone model.}
\label{fig:vin-initial-feature}
\end{subfigure}
\begin{subfigure}[b]{0.475\columnwidth}
\includegraphics[width=\textwidth]{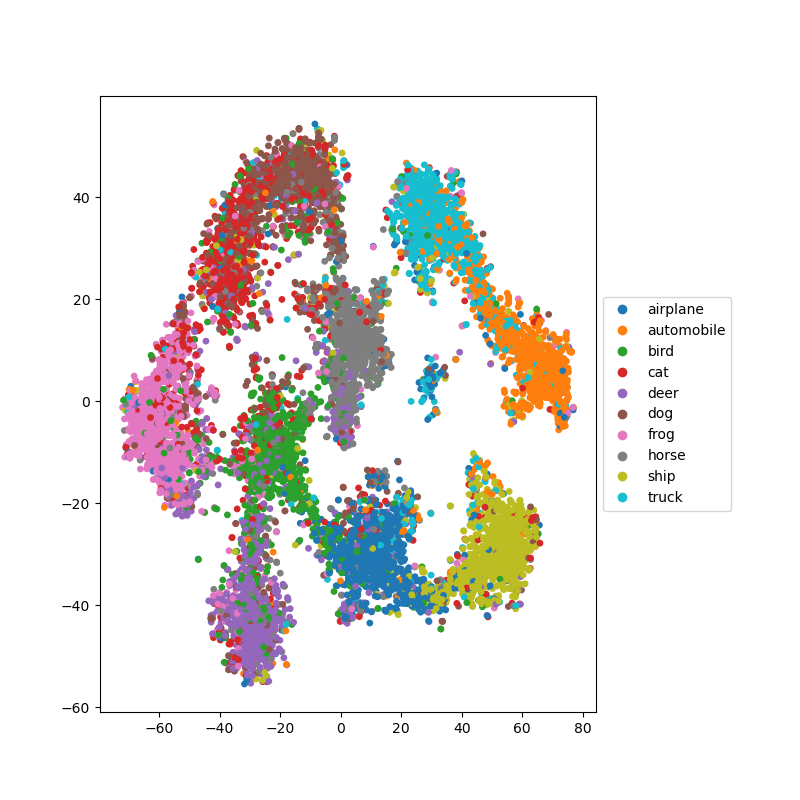}
\caption{t-SNE visualization of the learned feature of ViN backbone model with label smoothing.}
\label{fig:vin-smooth-feature}
\end{subfigure}

\begin{subfigure}[b]{0.475\columnwidth}
\includegraphics[width=\textwidth]{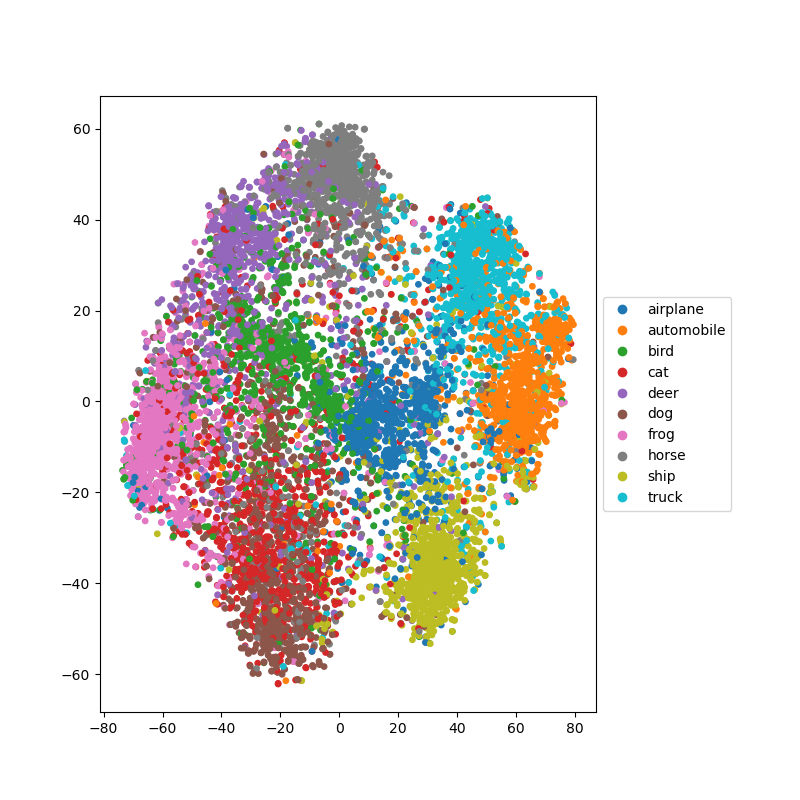}
\caption{t-SNE visualization of the learned feature of ViN backbone model with SEAL regularization.}
\label{fig:vin-stw-boosted-feature}
\end{subfigure}
\begin{subfigure}[b]{0.475\columnwidth}
\includegraphics[width=\textwidth]{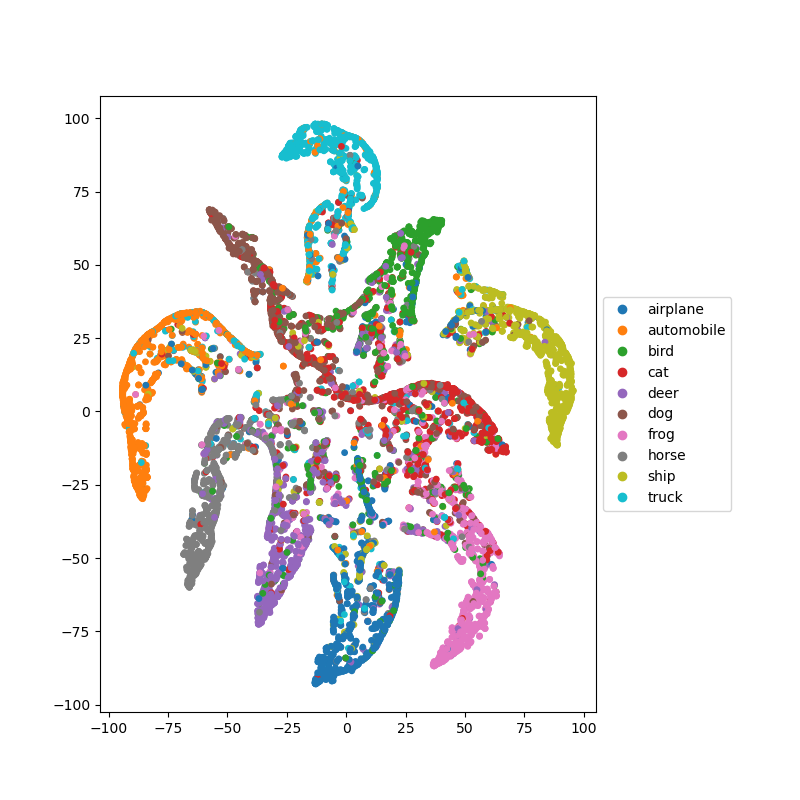}
\caption{t-SNE visualization of the learned probability of ViN model with SEAL regularization. The similarity metric is defined by relaxed Tree-Wasserstein distance.}
\label{fig:vin-stw-prob}
\end{subfigure}
\caption{Comparison of t-SNE visualizations of features and probabilities learned by the ViN model with different regularization methods. (a) shows the initial feature of the ViN backbone model. (b) shows the learned feature with label smoothing. (c) shows the learned feature with SEAL regularization. (d) shows the learned probability with SEAL regularization using relaxed Tree-Wasserstein distance as the similarity metric.}
\label{fig:vin-regularization-comparison}
\end{figure}

We are also interested in t-SNE visualization of the learned backbone feature, Fig. \ref{fig:vin-initial-feature} is the initial ViN feature, and Fig. \ref{fig:vin-smooth-feature} ViN with label smoothing $\alpha=0.1$, Fig. \ref{fig:vin-stw-boosted-feature} is the learned ViN feature with SEAL regularization. Fig. \ref{fig:vin-stw-prob} is slightly different, and we use the learned relaxed Tree-Wasserstein distance on probability space as the similarity metric.

As for the k-nearest neighbors (kNN) task, we choose the best $k$ for each subtask respectively. We summarize the result in Table \ref{tab:vin-knn}.
Note relaxed Tree-Wasserstein distance is a well-defined metric, so we also calculate the relaxed Tree-Wasserstein distance on probability space to do the knn task. The tree we used is plotted in Fig. \ref{vin-tree}. The tree shows some semantic relations between classes, as semantic closer classes have smaller tree distances.

\begin{table*}[t]
\caption{\label{tab:vin-knn} ViN best knn classification}
\centering
\begin{tabular}{llll}
\toprule
initial feature & smooth feature   & SEAL feature & prob space hard treedis        \\
\midrule
0.6725 & 0.6818  & 0.6831 & 0.8094  \\
\bottomrule
\end{tabular}
\end{table*}

\subsection{ResNet18 backbone}

\begin{table*}[t]
\caption{\label{tab:resnet-knn} ResNet18 best knn classification}
\centering
\begin{tabular}{llll}
\toprule
initial feature & smooth feature   & SEAL feature & prob space hard treedis        \\
\midrule
0.9542 & 0.9551  & 0.9572 & 0.9574  \\\bottomrule
\end{tabular}
\end{table*}

Inspired by paper NBDT, on CIFAR10 we train 200 epochs, and the origin (trained by cross-entropy loss) accuracy is 95.42\%. With label smoothing, the accuracy is 95.54\%, while with SEAL regularization, the accuracy is 95.75\%. In the above experiments, we train the first 180 epochs using the same loss as the initial and turn the loss to label smoothing or RTW respectively.

\begin{figure}[htb]
\centering
\begin{subfigure}[t]{0.49\linewidth}
\centering
\includegraphics[width=\linewidth]{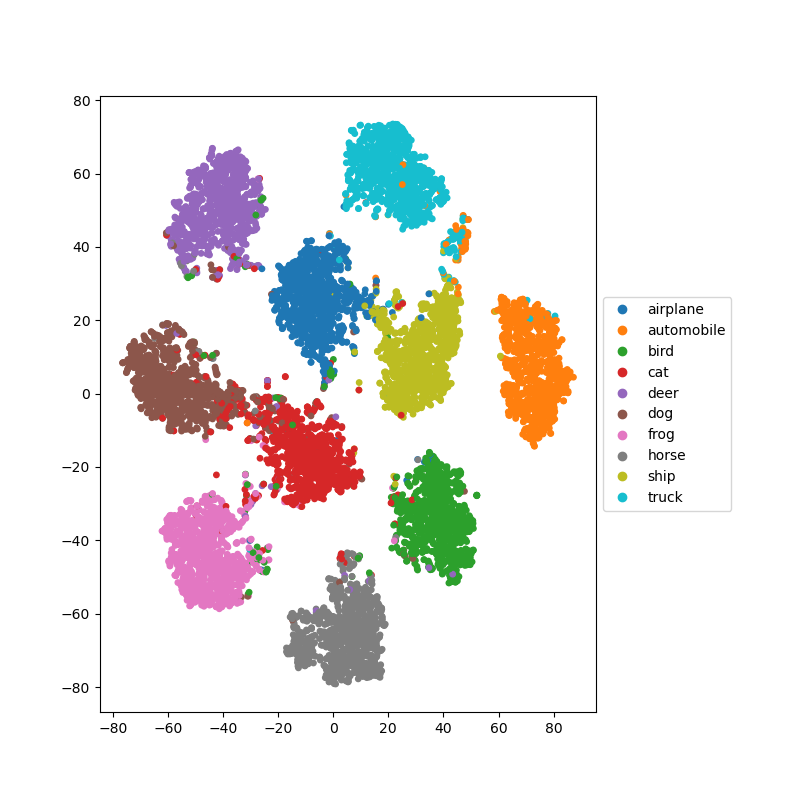}
\caption{t-SNE visualization of the initial feature of ResNet18 backbone model}
\label{fig:resnet-initial-feature}
\end{subfigure}
\hfill
\begin{subfigure}[t]{0.49\linewidth}
\centering
\includegraphics[width=\linewidth]{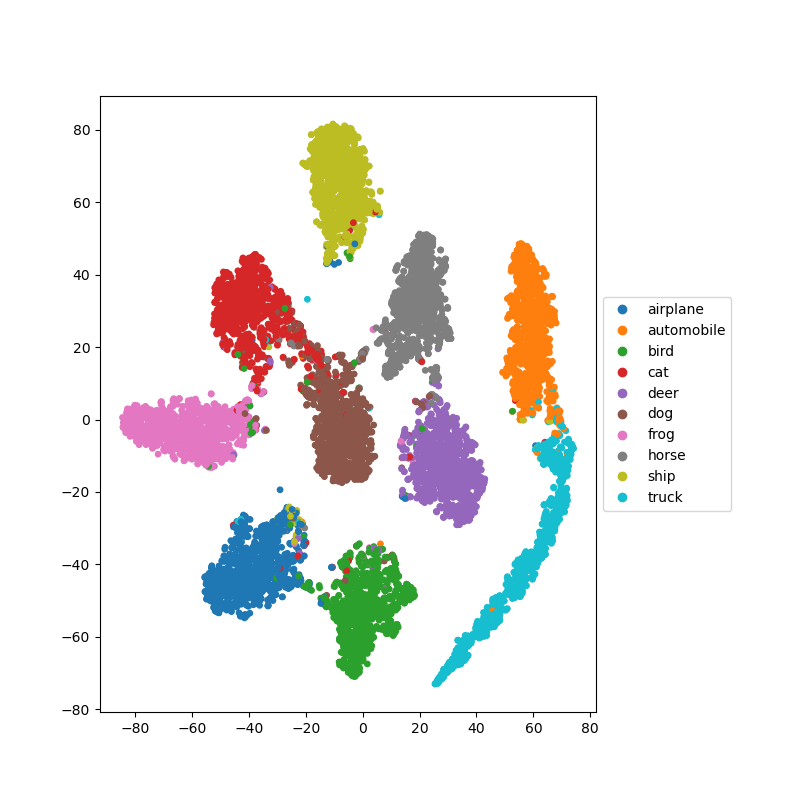}
\caption{t-SNE visualization of the learned feature of ResNet18 backbone model with label smoothing}
\label{fig:resnet-smooth-feature}
\end{subfigure}
\vspace{0.5cm}

\begin{subfigure}[t]{0.49\linewidth}
    \centering
    \includegraphics[width=\linewidth]{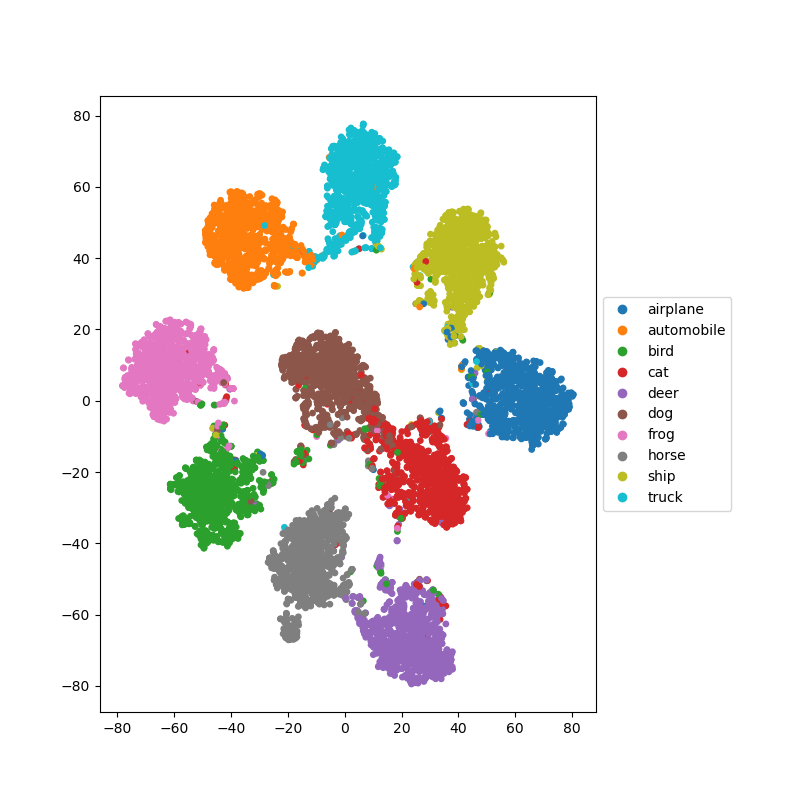}
    \caption{t-SNE visualization of the learned feature of ResNet18 backbone model with SEAL regularization}
    \label{fig:resnet-stw-feature}
\end{subfigure}
\hfill
\begin{subfigure}[t]{0.49\linewidth}
    \centering
    \includegraphics[width=\linewidth]{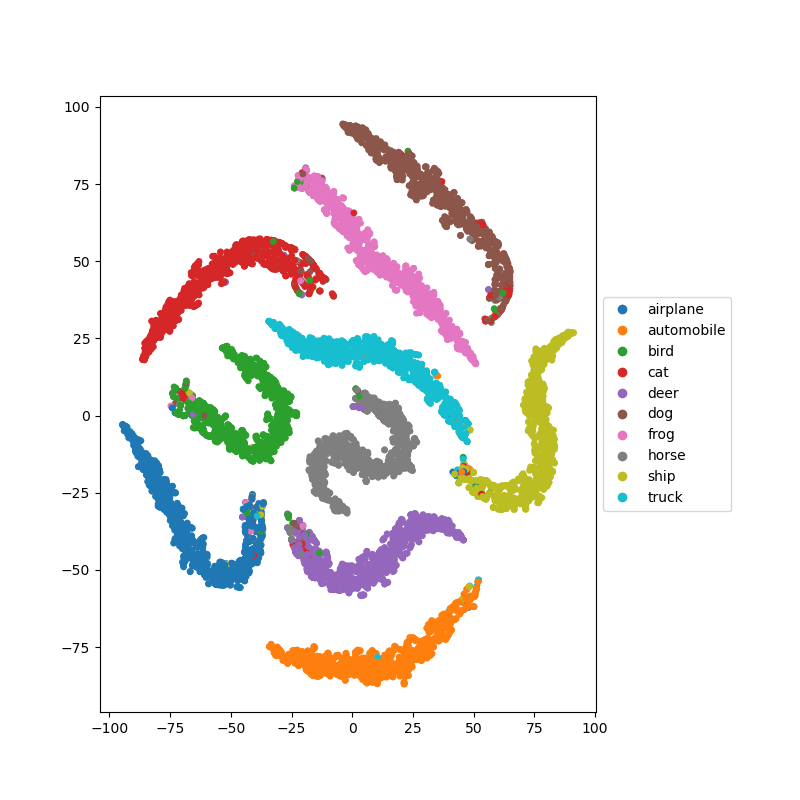}
    \caption{t-SNE visualization of the learned probability of ResNet18 backbone model with SEAL regularization. Similarity metric is defined by relaxed Tree-Wasserstein distance.}
    \label{fig:resnet-stw-prob}
\end{subfigure}
\caption{Visualizations of ResNet18 backbone model features with different regularization methods.}
\label{fig:resnet-features}
\end{figure}

We are also interested in t-SNE visualization of the learned backbone feature, Fig. \ref{fig:resnet-initial-feature} is the initial VIN feature, and Fig. \ref{fig:resnet-smooth-feature} VIN with label smoothing $\alpha=0.1$, Fig. \ref{fig:resnet-stw-feature} is the SEAL boosted VIN feature. Fig. \ref{fig:resnet-stw-prob} is slightly different, and we use the learned tree distance on probability space as the similarity criteria.

\begin{figure}[htb]
\centering
\begin{subfigure}[b]{0.49\columnwidth}
\includegraphics[width=\linewidth]{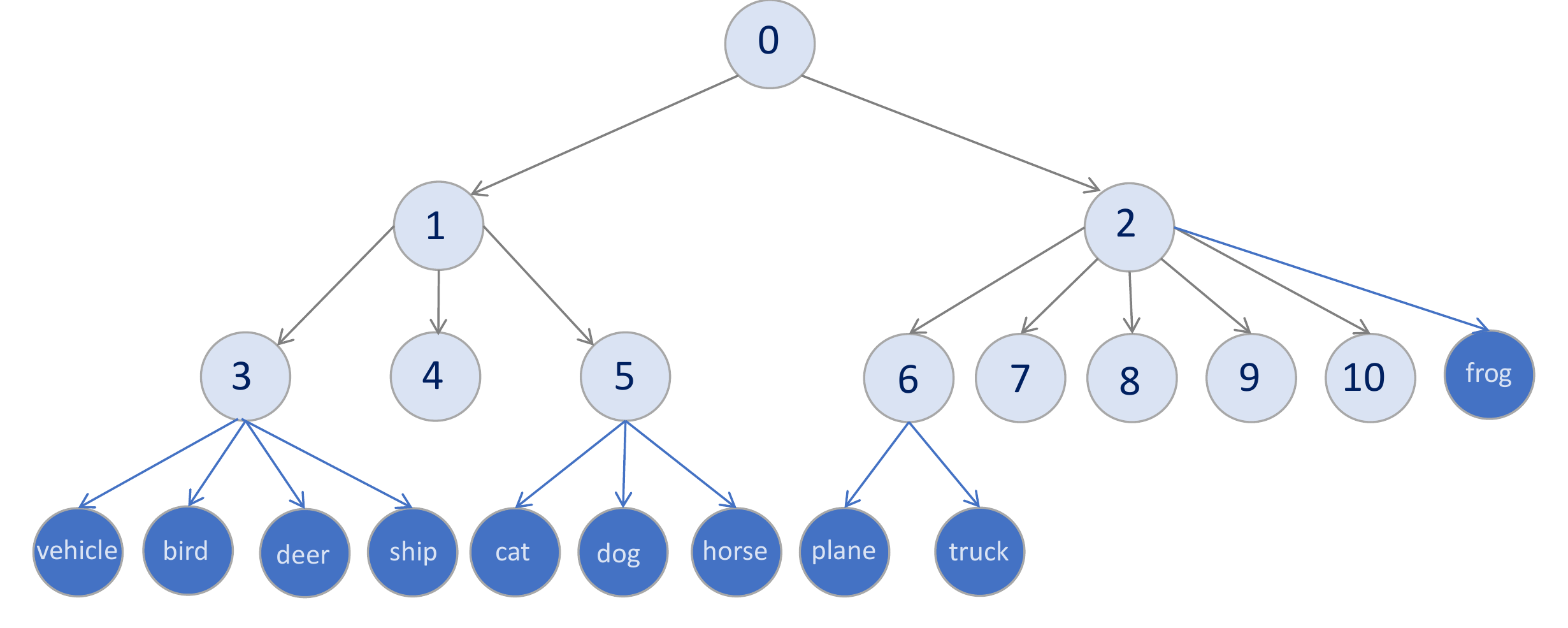}
\caption{ViN Tree}
\label{vin-tree}
\end{subfigure}
\begin{subfigure}[b]{0.49\columnwidth}
\includegraphics[width=\linewidth]{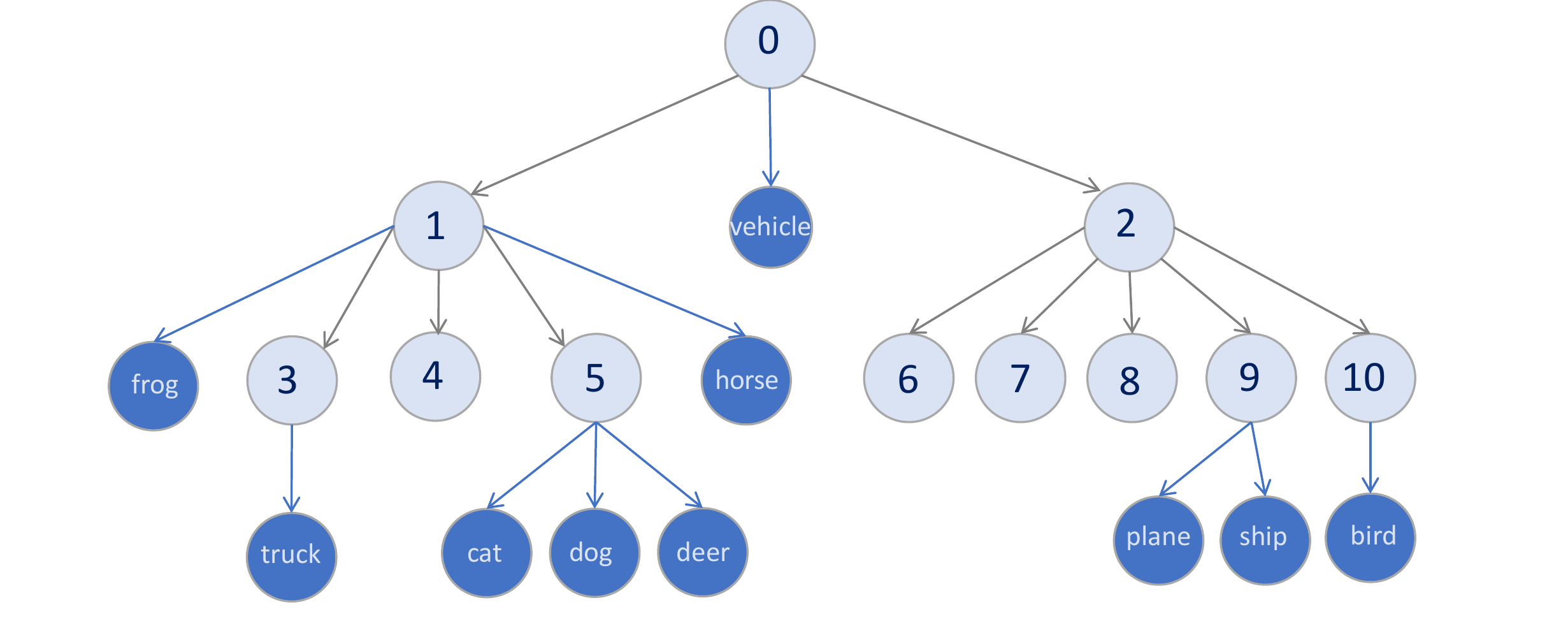}
\caption{ResNet Tree}
\label{resnet-tree}
\end{subfigure}
\caption{Comparison of ViN Tree and ResNet Tree}
\label{fig:comparison}
\end{figure}

As for the k-nearest neighbors (kNN) task, we summarize the result in Table \ref{tab:resnet-knn}. Note relaxed Tree-Wasserstein distance is a well-defined metric, so we also calculate the relaxed Tree-Wasserstein distance on probability space to do the knn task. The tree we used is plotted in Figure \ref{resnet-tree}. The tree shows some semantic relations between classes, as semantic closer classes have smaller relaxed Tree-Wasserstein distances.

Note that in the original paper NBDT, the initial accuracy is 94.97\%, their method gets 94.82\%. Their initial accuracy is slightly lower than ours may be due to the number of epochs they run being smaller.

\end{document}